\documentclass[letterpaper]{article} 
\usepackage{aaai2027} 
\usepackage[hyphens]{url}            
\usepackage{graphicx}                
\usepackage{natbib}                  
\usepackage{caption}                 
\usepackage[utf8]{inputenc} 
\usepackage[T1]{fontenc}    
\usepackage{booktabs}       
\usepackage{amsfonts}       
\usepackage{nicefrac}       
\usepackage{microtype}      
\usepackage[table]{xcolor} 
\usepackage{amsmath}
\usepackage{booktabs}
\usepackage{longtable}
\usepackage{array}
\usepackage{placeins}
\usepackage{multirow} 
\usepackage{subcaption}
\usepackage{adjustbox}

\newcolumntype{L}[1]{>{\raggedright\arraybackslash}p{#1}}
\usepackage{tabularx} 
\usepackage[most]{tcolorbox}
\tcbuselibrary{breakable}
\usepackage{tabularx}
\usepackage{booktabs}
\usepackage{array}
\usepackage{longtable}

\usepackage{booktabs}
\usepackage{array}
\usepackage{makecell}
\usepackage{longtable}
\usepackage{etoolbox}

\newcolumntype{Y}{>{\raggedright\arraybackslash}X}

\renewcommand{\arraystretch}{0.94}

\definecolor{cvprblue}{rgb}{0.21,0.49,0.74}

\usepackage{xcolor}

\usepackage{graphicx}
\usepackage{xcolor}
\usepackage{caption} 

\usepackage{cuted}

\definecolor{tokbg}{HTML}{FFE08A}

\newcommand{\mixedcomponent}[6]{%
  \noindent
  \begin{minipage}{\linewidth}
    \hrule height 0.4pt
    \vspace{3pt}

    \begin{tabular*}{\linewidth}{%
      @{\extracolsep{\fill}}lr@{}}
      {\footnotesize\textbf{#1}}
      &
      {\footnotesize\color{black!65}%
       Layer #2 \enspace$\vert$\enspace \#\,#3}
    \end{tabular*}

    \vspace{1pt}

    {\footnotesize
      \raggedright
      \textbf{Label:} #6\par
    }

    \vspace{4pt}

    \begin{minipage}[t]{0.40\linewidth}
      \vspace{0pt}
      \centering
      \includegraphics[width=\linewidth]{#4}
    \end{minipage}%
    \hfill
    \begin{minipage}[t]{0.56\linewidth}
      \vspace{0pt}
      \raggedright
      #5
    \end{minipage}

    \vspace{4pt}
  \end{minipage}
  \par
}

\title{Through the LENS: Local Geometric Decomposition of Vision-Language Model Representations}

\author{
    Shalom Kachko\textsuperscript{\rm 1,*},
    Raz Lapid\textsuperscript{\rm 2,*},
    Margarita Vald\textsuperscript{\rm 2},
    Almog Dubin\textsuperscript{\rm 3},
    Moshe Sipper\textsuperscript{\rm 1}
}

\affiliations{
    \textsuperscript{\rm 1}Ben-Gurion University of the Negev\\
    \textsuperscript{\rm 2}Intuit Inc.\\
    \textsuperscript{\rm 3}Deepkeep\\
    kachko.shulik@gmail.com,
    raz\_lapid@intuit.com,
    margarita.vald@cs.tau.ac.il,
    almog@deepkeep.ai,
    sipper@bgu.ac.il
}

\usepackage{amsmath,amsfonts,bm,upgreek}

\def\eqref#1{eq.~(\ref{#1})}

\def\1{\bm{1}}

\DeclareMathAlphabet{\mathsfit}{\encodingdefault}{\sfdefault}{m}{sl}
\SetMathAlphabet{\mathsfit}{bold}{\encodingdefault}{\sfdefault}{bx}{n}

\usepackage[capitalize,noabbrev]{cleveref}

\begin{document}

\maketitle

\begingroup
\renewcommand{\thefootnote}{*}
\footnotetext{These authors contributed equally.}
\endgroup

\begin{abstract}
Vision-language models (VLMs) process image patches and text tokens in a shared residual stream, but the local geometry through which the two modalities interact remains poorly understood. Most interpretability methods identify global linear directions, which may miss representations that are globally high-dimensional but locally low-dimensional. We introduce \textbf{LENS} (\textbf{L}ocal \textbf{E}xplanation of \textbf{N}eighborhood \textbf{S}ubspaces), a method that decomposes VLM activations into local low-rank Gaussian neighborhoods using a Mixture of Factor Analyzers. Applied to LLaVA-1.5-7B and Qwen3-VL-8B, LENS reveals distinct depth-dependent fusion trajectories consistent with each model's fusion mechanism: LLaVA progressively mixes modalities at later layers, whereas Qwen3-VL mixes them early, partially re-segregates them, and recombines them near the output. An automated multimodal labeling pipeline assigns concise semantic descriptions to these neighborhoods. Interpolating activations toward neighborhood centroids causally redirects generation within and across modalities and outperforms difference-in-means and VL-SAE in most evaluated conditions; in one LLaVA vision-to-vision setting, MFA achieves 5.7 times the VL-SAE score. Human evaluation finds MFA steering competitive with prompting and substantially stronger than the other intervention baselines. Finally, the MFA coefficient space improves Qwen3-VL image-to-rendered-text retrieval at the deepest evaluated layer from 14.9\% to 48.6\% R@1. Ablations show that the reported fusion trajectories are stable across component counts, local ranks, and modality-purity thresholds. These results support local geometric neighborhoods as useful interpretable and causal units for analyzing cross-modal representations in the evaluated VLMs.
\end{abstract}    
\section{Introduction}
\label{sec:intro}

Vision-language models (VLMs) jointly process images and text for
captioning, visual question answering, instruction following, and visual
reasoning
\citep{radford2021learning,alayrac2022flamingo,li2023blip,
liu2023visual,zhang2023multimodal}.
Their growing use in high-stakes applications-including medical diagnosis, remote-sensing analysis, and industrial quality control 
\citep{xiang2025vision,hu2025rsgpt,
yang2025surveyfoundationmodelbasedindustrialdefect}, motivates understanding how they combine visual and linguistic evidence. Yet cross-modal fusion remains poorly understood
\citep{lin2025survey,zhang2025cross}. Most VLMs project visual patches and text tokens into a shared residual stream, where nearby representations may encode visual content, linguistic context, or interactions between the two \citep{elhage2022toy}. We still lack tools to identify where modalities meet, characterize these regions, and test their causal control over model behavior.

Mechanistic interpretability commonly addresses polysemanticity - the phenomenon  where individual neurons encode multiple disparate concepts in superposition.  This is typically achieved using Sparse Autoencoders (SAEs), which represent activations as sparse combinations of learned directions \citep{bricken2023monosemanticity,Cunningham2024sparse}.
In multimodal models, SAEs have been applied to vision-encoder
representations
\citep{hugofry_towards_multimodal_interpretability,pach2025sparse},
aligned across vision architectures
\citep{thasarathan2025universal}, and trained on internal representations of large language models (LLMs) for interpretation and steering
\citep{zhang2025large}.
VL-SAE further learns a unified concept space for visual and linguistic
representations \citep{shen2026vlsae}, while complementary work traces
cross-modal information flow inside VLMs \citep{zhang2025cross}.
Despite their differences, these approaches generally describe activations using globally shared linear features or directions, potentially missing representations that are
high-dimensional globally but low-dimensional locally
\citep{lee2025shared,saglam_llm_separable_representations}. Recent work in unimodal language models \citet{shafran2026directions} instead uses a Mixture of Factor Analyzers (MFA) \citep{ghahramani1996em} to represent activations as Gaussian neighborhoods with locally low-dimensional structure.

\textbf{From unimodal to cross-modal geometry.}
Extending this local-geometric approach to VLMs is nontrivial because image and text representations enter through different pathways and may occupy geometrically separated regions, as documented by studies of the modality gap \citep{liang2022mind}. A multimodal decomposition must accommodate heterogeneous token distributions, interpret neighborhoods containing both image patches and text contexts, and test whether their semantics can be accessed causally across modalities.

We introduce \textbf{LENS}
(\textbf{L}ocal \textbf{E}xplanation of \textbf{N}eighborhood
\textbf{S}ubspaces), which decomposes token-level activations in the
fused VLM residual stream into Gaussian neighborhoods with local
low-rank geometry.
LENS identifies where modalities share representation space, labels
these neighborhoods from associated image patches and text contexts,
and uses them as causal steering targets and as shared coordinates for cross-modal comparison.
Applied to LLaVA-1.5-7B \citep{liu2023visual} and
Qwen3-VL-8B-Instruct \citep{yang2025qwen3technicalreport}, LENS reveals distinct depth-dependent fusion patterns associated with the two architectures: progressive late fusion in LLaVA
and early fusion, partial re-segregation, and late recombination in
Qwen3-VL.
Human evaluation supports the automated steering results: MFA achieves the highest median human-evaluation score overall and substantially outperforms VL-SAE and DiffMeans, although its performance relative to prompting varies by architecture. Human ratings are also moderately correlated with the VLM-judge scores (Spearman's $\rho=0.727$ Pearson's $r=0.747$), supporting the validity of the automated evaluation while motivating independent human
validation.

Our contributions are:

\begin{itemize}
    \item \textbf{Local-geometric decomposition and interpretation.}
    We introduce an MFA-based decomposition of fused VLM residual streams, together with a pipeline for labeling neighborhoods from
    associated image patches and text contexts.

    \item \textbf{Architecture-dependent fusion geometry.}
    LENS reveals progressive late fusion in LLaVA and early,
    non-monotonic fusion in Qwen3-VL. These depth-dependent patterns remain stable across the evaluated component counts, local ranks, and modality-purity thresholds.

    \item \textbf{Causal and functional validation.}
    MFA centroid steering outperforms DiffMeans and VL-SAE in most early- and middle-layer conditions and enables both within- and cross-modal control. Human evaluation supports these gains. In addition, MFA coefficient representations improve retrieval between natural images, rendered class names, and textual class labels over raw activations in several settings.
\end{itemize}
\section{Related Work and Background}
\label{sec:related}

\subsection{Dictionary Learning and VLM Interpretability}
Mechanistic interpretability commonly addresses polysemantic representations using dictionary-learning methods, particularly Sparse Autoencoders (SAEs)~\citep{bricken2023monosemanticity,templeton2024scaling}. SAEs map model activations into a sparsely activating latent space and reconstruct them from a learned dictionary of feature directions. In VLMs, recent work applies SAEs to unimodal vision encoders \citep{pach2025sparse} aligns learned features across vision architectures \citep{thasarathan2025universal}, and interprets internal representations of large multimodal models \citep{zhang2025large}. Most closely related to our work, VL-SAE learns a unified concept space for visual and linguistic representations through cascaded decoders \citep{shen2026vlsae}.

However, these methods rely on globally shared feature directions, which can yield non-atomic features~\citep{chanin2024absorption} and miss structure that is low-dimensional only locally~\citep{saglam_llm_separable_representations}. Prior work also identifies a global \emph{modality gap} between image and text embeddings~\citep{liang2022mind,zhang2025cross}, but how this separation is bridged within fused VLM residual streams remains unclear.

\subsection{Local Geometry and Factor Analysis}
Recent work suggests transformer representations share global structure while retaining model-specific local geometry~\citep{lee2025shared}. Mixtures of Factor Analyzers (MFA)~\citep{ghahramani1996em} model this structure as $K$ local Gaussian neighborhoods with low-rank subspaces. \citet{shafran2026directions} showed that such components support concept localization and causal steering in language models.

\textbf{Positioning.} LENS extends MFA to VLMs by addressing three challenges: identifying mixed-modality regions despite cross-modal separation; grounding components in both image patches and text contexts; and enabling modality-specific interventions. LENS therefore treats local low-rank neighborhoods as interpretable and steerable units for analyzing cross-modal fusion.
\section{Local Explanation of Neighborhood Subspaces}
\label{sec:method}

\subsection{Preliminaries}

\paragraph{VLM activations.} Let $\mathcal{M}$ denote a VLM with $L$ layers, and let $\ell \in \{1,\ldots,L\}$ be a target residual-stream layer. For a multimodal input containing visual patches and text tokens, we extract the hidden state at each selected token or patch position and treat it as an activation vector $\mathbf{x}\in\mathbb{R}^{d}$. Processing a corpus of $N$ multimodal prompts yields multiple
token- and patch-level activations per prompt. We denote the
resulting activation dataset at layer $\ell$ by
\[
X_\ell=\{x_i\}_{i=1}^{S_M},
\]
where $S_M$ is the total number of sampled activations for
model $M$ across the corpus. Unlike $N$, $S_M$ depends on the
model architecture, input resolution, tokenizer, and sampling
scheme. Each activation is tagged with its source modality and
contextual metadata.

\paragraph{Mixture of Factor Analyzers (MFA).} MFA~\citep{ghahramani1996em} models an activation space as $K$ local low-rank Gaussian neighborhoods. A discrete latent $\omega \in \{1,\ldots,K\}$ assigns each sample to a component; conditioned on $\omega = k$,
\begin{equation}
\mathbf{x} = \boldsymbol{\mu}_k + \mathbf{W}_k \mathbf{z}_k + \boldsymbol{\epsilon},
\quad \text{with } \mathbf{z}_k \sim \mathcal{N}(\mathbf{0}, \mathbf{I}_q), \quad \boldsymbol{\epsilon} \sim \mathcal{N}(\mathbf{0}, \boldsymbol{\Psi}),
\end{equation}
where $\mathbf{x} \in \mathbb{R}^d$ is the observed activation; $\boldsymbol{\mu}_k \in \mathbb{R}^d$ is the center of the $k$-th neighborhood; $\mathbf{W}_k \in \mathbb{R}^{d \times q}$ is the factor-loading matrix, whose columns span the orientation of the component's local low-dimensional subspace; $q \ll d$ is the local latent rank per layer; $\mathbf{z}_k \in \mathbb{R}^{q}$ is the latent factor with a standard-normal prior; and $\boldsymbol{\epsilon} \in \mathbb{R}^d$ is coordinate-wise-independent zero-mean Gaussian noise with diagonal covariance $\boldsymbol{\Psi} \in \mathbb{R}^{d \times d}$. The resulting component covariance is $\mathbf{C}_k = \mathbf{W}_k \mathbf{W}_k^\top + \boldsymbol{\Psi}$. The marginal density is the weighted sum
\begin{equation}
p(\mathbf{x}) = \sum_{k=1}^{K} \pi_k\, \mathcal{N}(\mathbf{x} \mid \boldsymbol{\mu}_k, \mathbf{C}_k),
\end{equation}
where $\pi_k \ge 0$ is the mixture weight of component $k$, with $\sum_{k=1}^{K} \pi_k = 1$.
Conditioned on component $k$, an activation is represented by
the neighborhood centroid $\mu_k$ and a low-rank
within-neighborhood offset $W_k z_k$. The posterior probability that component $k$ generated $\mathbf{x}$--its \emph{responsibility}--is
\begin{equation}
\label{eq:responsibility}
R_k(\mathbf{x}) = \frac{\pi_k\, \mathcal{N}(\mathbf{x} \mid \boldsymbol{\mu}_k, \mathbf{C}_k)}{\sum_{j} \pi_j\, \mathcal{N}(\mathbf{x} \mid \boldsymbol{\mu}_j, \mathbf{C}_j)}.
\end{equation}
Conditioned on $\omega{=}k$, the posterior over the latent factors is itself Gaussian; its mean gives the \emph{local coordinate} of $\mathbf{x}$ within component $k$'s subspace,
\begin{equation}
\label{eq:posterior-mean}
\hat{\mathbf{z}}_k(\mathbf{x}) = \mathbf{W}_k^\top \mathbf{C}_k^{-1}\,(\mathbf{x} - \boldsymbol{\mu}_k) \in \mathbb{R}^q .
\end{equation}
Responsibilities $R_k$ form a soft partition of activation space and local coordinates $\hat{\mathbf{z}}_k$ locate a sample within each region; together they underlie our labeling, steering, and retrieval procedures.

\subsection{The LENS Pipeline}\label{sec:lens_pipeline}

\begin{figure*}[ht!]
\centering
  \includegraphics[scale=0.3]{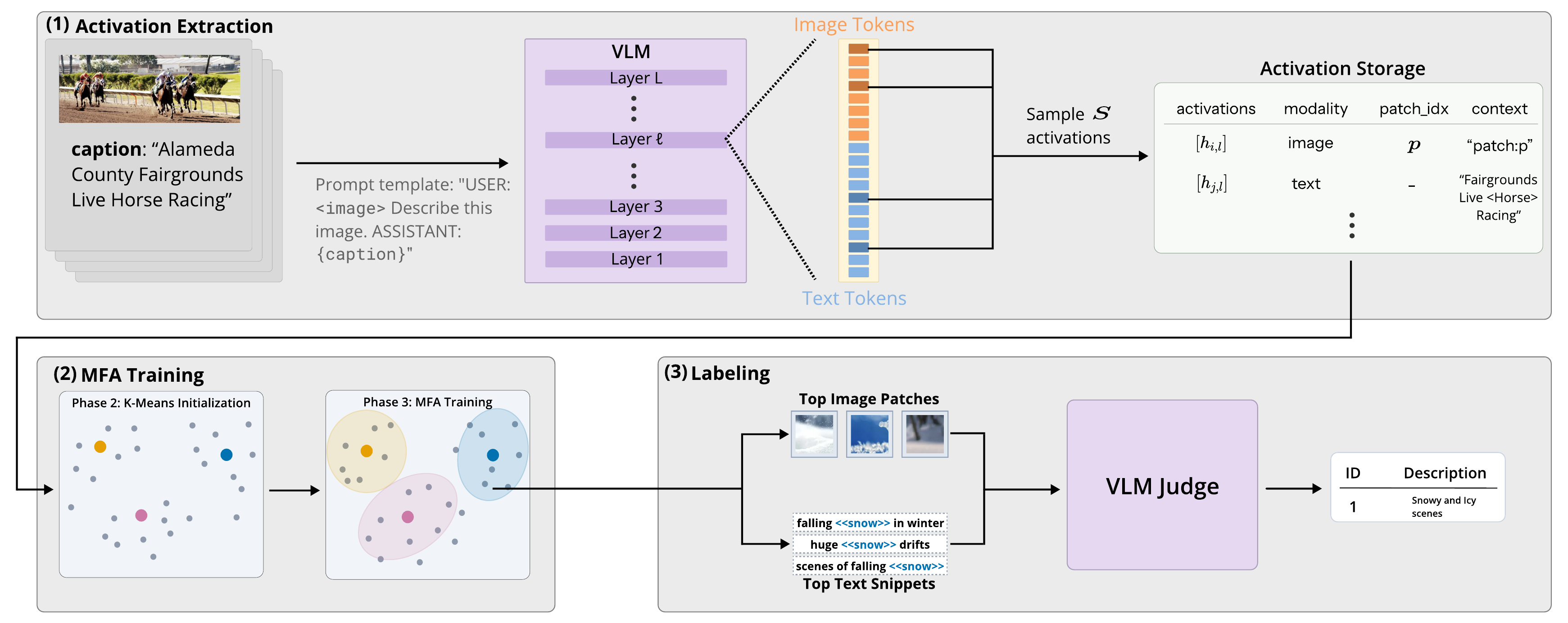}
    \caption{
    Activation extraction and labeling pipeline. 
    (1) Image-caption pairs are processed by a VLM to extract and store intermediate multimodal activations.
    (2) A Mixture of Factor Analyzers (MFA) is trained on the sampled activation space.
    (3) learned components are interpreted by a VLM Judge, which analyzes the component's top-activating image patches and text snippets and generates a concise, human-readable semantic description.
    }
    \label{fig:experiments}
\end{figure*}

LENS comprises three stages: (i) multimodal activation extraction, (ii) MFA training, and (iii) automated cross-modal labeling of the MFA components. Figure~\ref{fig:experiments} illustrates the extraction and deferred patch-labeling pipeline.

\subsubsection{Stage 1: Multimodal activation extraction.}\label{subsec:extraction}

Given a dataset $\mathcal{D} = \{(\mathbf{I}_n, \mathbf{c}_n)\}_{n=1}^N$ of image-caption pairs, we extract per-token residual-stream activations from target intermediate layers. To guarantee deterministic patch geometry across architectures that handle dynamic input resolutions, we pin the model to a fixed square 
input resolution. Specifically:

\begin{itemize}
    \item \textit{Stratified modality sampling.} Each token position is classified with the class $m$  where $m{=}\texttt{img}$ for image-placeholder positions and $m{=}\texttt{txt}$ for non-special, non-padding positions. Image tokens typically outnumber text tokens by an order of magnitude, so a naive corpus would be vision-dominated and obscure cross-modal regions. We therefore sample $\min(|\mathcal{I}|, |\mathcal{T}|, h_{\max})$ positions \emph{per modality per pair}, where $|\mathcal{I}|$ is the number of image tokens, $|\mathcal{T}|$ is the number of text tokens, and $h_{\max}=10$ is a per-pair budget. This yields a corpus with comparable mass on each modality.

    \item \textit{Metadata for downstream labeling.} Each extracted activation $\mathbf{x}$ is stored with its modality flag $m$. For image tokens $\mathcal{I}$ we additionally record a spatial patch index $p$ (row, column in the vision encoder grid); for text tokens $\mathcal{T}$ we record a windowed token context with the target token marked. This metadata enables deferred multimodal labeling (Section~\ref{sec:labeling}) without re-running the VLM at extraction time.

\end{itemize}

\subsubsection{Stage 2: MFA training.}
\label{sec:fitting}

We train a separate MFA independently at each target layer $\ell$, since activation geometry changes with depth. Unless otherwise stated, we use the same component count $K$ and local latent rank $q$ across target layers; $K$ and $q$ are the principal hyperparameters, specified in Section~\ref{sec:exp_setup} and ablated in Appendix~\ref{app:ablation}. 

Direct $K$-means initialization in the full residual dimension is prohibitive, so we follow \citet{shafran2026directions}: cluster activations in a $d_{\text{proj}}{=}256$-dimensional random orthonormal projection, then initialize each full-space centroid $\boldsymbol{\mu}_k$ as the mean of the original activations assigned to that projected cluster. Loadings $\{\mathbf{W}_k\}_{k=1}^K$ are initialized with entries drawn i.i.d. from $\mathcal{N}(0,1)$, the shared diagonal noise is initialized as $\boldsymbol{\Psi} = \mathbf{I}_d$, and the mixing weights are initialized uniformly as $\pi_k = \frac{1}{K}$ for all $k$.

Training minimizes the negative log-likelihood
\begin{equation}
\mathcal{L}(\theta) = -\frac{1}{B}\sum_{n=1}^{B} \log \left( \sum_{k=1}^{K} \pi_k\, \mathcal{N}(\mathbf{x}_n \mid \boldsymbol{\mu}_k, \mathbf{C}_k)\right),
\end{equation}
over $\theta = \{\pi_k, \boldsymbol{\mu}_k, \mathbf{W}_k, \boldsymbol{\Psi}\}_{k=1}^{K}$ using Adam optimizer \citep{kingma2017adammethodstochasticoptimization}, with learning rate $10^{-3}$, gradient clipping at norm $1.0$ and batch size $B=1024$.

\subsubsection{Stage 3: Automated cross-modal labeling.}
\label{sec:labeling}

A key methodological challenge in extending MFA to VLMs is that some components are \emph{mixed}: their high-responsibility or hard-assigned activations include a substantial fraction of both visual patches and text tokens. Producing a label that reflects what the model represents across modalities therefore requires more than the text-only protocol of \citet{shafran2026directions}. We keep the main procedure concise here and provide full details in Appendix~\ref{app:prompt_labeling}.
\begin{enumerate} \setlength{\parskip}{0.5pt}

    \item \emph{Component modality and top activations.}
    We select a subset of $K'$ MFA components. We assign each
    activation to the component with maximum responsibility and
    compute each component's modality purity over all hard-assigned
    activations. Given a purity threshold $\tau$ (set to $\tau=0.8$
    in our experiments), a component is text-dominated or
    vision-dominated if the corresponding modality accounts for at
    least a fraction $\tau$ of its assignments, and mixed otherwise. Separately, to obtain examples for labeling, we stream the
    activation dataset through the trained MFA and retain the top-$t$
    activations for each selected component by responsibility
    $R_k(\cdot)$ (Equation~3), using a min-heap.

    \item \emph{Visual patch extraction.} 
    For each retained image activation, we locate its source image, resize it to the model's expected resolution, and crop a fixed-size window centered on the patch's spatial coordinate, isolating the image region responsible for its high responsibility score.

    \item \emph{Vision-language judge-based synthesis.} 
    For each selected component, we use a judge VLM to produce one concise natural-language description of the component. The judge is provided with contexts of the top text activations and the image patches cropped in \textbf{(2)}, both ordered by responsibility together with a strict system prompt detailed in Appendix \ref{app:prompt_labeling}. Our prompt spans both modalities, and hence extends prior automated-interpretability protocols~\citep{bills2023language,bricken2023monosemanticity} from the text-only setting to the cross-modal regime.
    

\end{enumerate}

\subsection{Causal Evaluation Interventions}
\label{sec:interventions}

The labeled decomposition produced by Section \ref{sec:lens_pipeline} is descriptive; we additionally test whether the recovered components are \emph{causal} handles on model behavior. We use two complementary interventions.

\paragraph{Scoped activation steering.} To test causal influence, we intervene during generation by interpolating hidden states at the target layer toward a component centroid:
\begin{equation}
\label{eq:steering}
\mathbf{x}' = (1-\alpha)\,\mathbf{x} + \alpha\,\boldsymbol{\mu}_k,
\qquad \mathbf{x} \in \mathcal{S},
\end{equation}
where $\mathcal{S}$ selects the intervention \emph{scope}--image activations only, text activations only, or both--and $\alpha \in [0, 1]$ controls intervention strength. The scope distinction is unique to multimodal models: unlike unimodal models, VLMs allow us to target visual or textual positions independently. This lets us ask not only \emph{whether} a component matters, but \emph{through which modality} it exerts its effect (Section~\ref{sec:exp_steering}).

\paragraph{Cross-modal coefficient-space retrieval.}
The MFA induces a shared coordinate system in which image and text
activations can be compared directly. The responsibility vector
\begin{equation}
\label{eq:responsibility-vector}
\mathbf{r}(\mathbf{x})
=
\big[R_1(\mathbf{x}),\ldots,R_K(\mathbf{x})\big]
\in [0,1]^K,
\qquad
\sum_{k=1}^{K} R_k(\mathbf{x}) = 1,
\end{equation}
encodes which Gaussian neighborhoods are active. To additionally encode
where an activation lies within each neighborhood, we weight each local
coordinate $\hat{\mathbf{z}}_k(\mathbf{x})$
(Equation~\ref{eq:posterior-mean}) by its responsibility and stack the
results:
\begin{equation}
\label{eq:b-of-x}
\mathbf{b}(\mathbf{x}) =
\begin{bmatrix}
R_1(\mathbf{x}) \\
R_1(\mathbf{x})\,\hat{\mathbf{z}}_1(\mathbf{x}) \\
\vdots \\
R_K(\mathbf{x}) \\
R_K(\mathbf{x})\,\hat{\mathbf{z}}_K(\mathbf{x})
\end{bmatrix}
\in \mathbb{R}^{K(1+q)}.
\end{equation}
Thus, $\mathbf{r}(\mathbf{x})$ encodes \emph{which} neighborhoods are
active, while $\mathbf{b}(\mathbf{x})$ also encodes \emph{where} within
them the activation lies. Both representations are defined in component
space and are therefore shared across modalities. We use cosine
similarity in $\mathbf{r}$- or $\mathbf{b}$-space for cross-modal
retrieval, without contrastive training
(Section~\ref{sec:exp_retrieval}).
\section{Experiments}
\label{sec:experiments}

We evaluate whether LENS (i) identifies local regions that mix visual
and textual evidence, (ii) provides causal handles for within- and
cross-modal steering, and (iii) exposes semantic alignment through
cross-modal retrieval.

\subsection{Experimental Setup}
\label{sec:exp_setup}
\paragraph{Models and data.}  
We study two open-source VLMs with distinct fusion architectures:
LLaVA-1.5-7B~\citep{liu2023visual}, which inserts projected visual
features at the input, and Qwen3-VL-8B-Instruct
\citep{yang2025qwen3technicalreport}, whose DeepStack design injects
multi-level visual features into several early layers
\citep{meng2024deepstack,yang2025qwen3technicalreport}.
To obtain broad coverage of naturally occurring cross-modal concepts, we use image-caption pairs from CC3M \citep{sharma-etal-2018-conceptual} and extract residual-stream activations at layers $8$, $16$, and $29$ for LLaVA and layers $9$, $18$, and $27$ for Qwen3. Fixed input resolutions of $336{\times}336$ and $448{\times}448$, respectively. Full implementation and hardware details are provided in
Appendix~\ref{app:implementation}.

\paragraph{Decomposition and evaluation.}
At each layer, we train an MFA with $K{=}8{,}192$ components and local rank
$q{=}10$ using the procedure in Section~\ref{sec:fitting}; an ablation
study over $K$ and $q$ is reported in Appendix~\ref{app:ablation}.
Geometry and retrieval use the full mixture. For automated labeling, we
randomly sample $1{,}000$ components, retain their $25$
highest-responsibility activations, and label them with
Qwen2.5-VL-7B-Instruct
\citep{bai2025qwen25vltechnicalreport}. 
Experiment-specific baselines and metrics are described below.

\subsection{ Cross-Modal Structure via Local Geometry}
\label{sec:exp_geometry}

\begin{figure*}[t]
  \centering
  \resizebox{1\textwidth}{!}{
  \definecolor{tokbg}{HTML}{FFE08A}
  \newcommand{\hltok}[1]{%
    \setlength{\fboxsep}{1.5pt}%
    \colorbox{tokbg}{\textbf{#1}}%
  }
  \newcommand{\compid}[3]{%
    {\footnotesize\color{black!70}%
    \begin{tabular}[c]{@{}c@{}}\textbf{#1}\\ Layer #2\\ \#\,#3\end{tabular}}}
  \newcommand{\tokstack}[1]{%
    {\footnotesize\renewcommand{\arraystretch}{1.2}%
    \begin{tabular}[c]{@{}l@{}}#1\end{tabular}}}
  \setlength{\tabcolsep}{8pt}
  \renewcommand{\arraystretch}{1.3}
  \begin{tabular}{@{} >{\centering\arraybackslash}m{0.12\textwidth} >{\centering\arraybackslash}m{0.18\textwidth} >{\centering\arraybackslash}m{0.28\textwidth} >{\raggedright\arraybackslash}m{0.32\textwidth} @{}}
    \toprule
    \multicolumn{1}{c}{\textbf{Component}} 
      & \multicolumn{1}{c}{\textbf{Image patches}}
      & \multicolumn{1}{c}{\textbf{Text tokens}}
      & \multicolumn{1}{c}{\textbf{Component label}} \\
    \midrule
    \compid{LLaVA-1.5}{29}{4982}
      & \begin{tabular}[c]{@{}c@{}}\includegraphics[width=\linewidth]{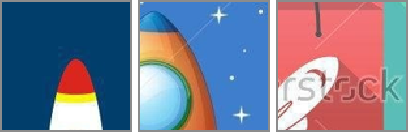}\end{tabular}
      & \tokstack{%
          strange \hltok{objects} and lights\\
          tennis player \hltok{launch}es his\\
          pop artist \hltok{launch}es her\\
          with ro\hltok{cket} in the\\
          model of a \hltok{module}}
      & Space exploration and launch activities are strongly activating themes. \\
    \midrule
    \compid{Qwen3-VL-8B}{27}{2532}
      & \begin{tabular}[c]{@{}c@{}}\includegraphics[width=\linewidth]{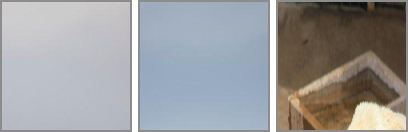}\end{tabular}
      & \tokstack{%
          p\hltok{olar} bear in\\
          baby \hltok{polar} bear enjoying\\
          these \hltok{polar} bears are\\
          card with a \hltok{polar}\\
          p\hltok{olar} bears , biological}
      & Polar bears are consistently activated across diverse visual and textual contexts. \\
    \bottomrule
  \end{tabular}
  }
  \caption{\textbf{Mixed components bind specific visual and textual content.}
    Two representative mixed components (LLaVA-1.5 layer~29 and Qwen3-VL-8B
    layer~27), each shown with its top max-activating image patches (left),
    text-token contexts with the responsible token highlighted (middle), label (right). Both activate on visual patches \emph{and}
    text tokens denoting the same concept.}
  \label{fig:mixed_components_main}
\end{figure*}

\begin{figure}[t]
    \centering
    \includegraphics[width=1\linewidth]{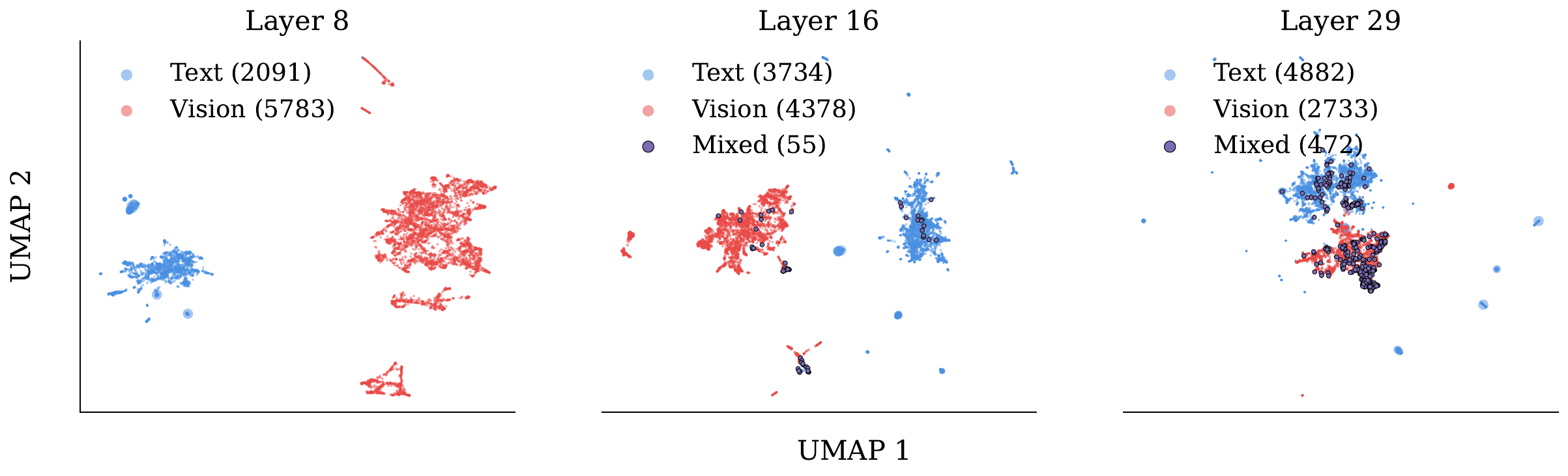}
    \includegraphics[width=1\linewidth]{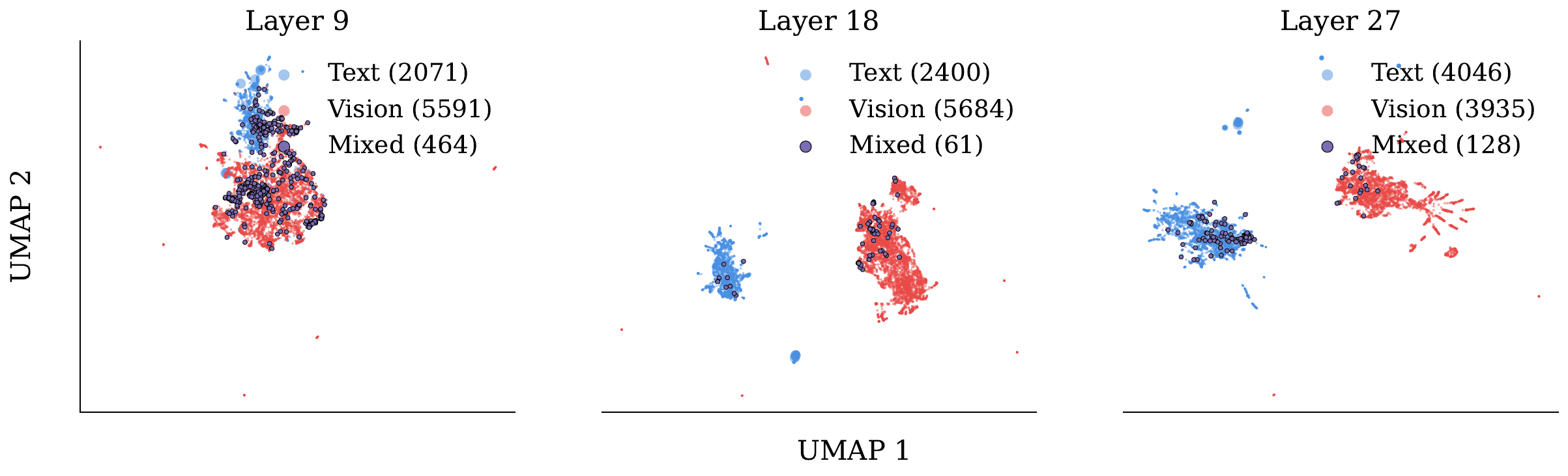}
    \caption{\textbf{Cross-modal fusion regimes across architectures.} \textbf{Top (LLaVA-1.5-7B):} \emph{Increasing late-layer mixing.} Mixed centroids accumulate monotonically with depth ($0 \to 55 \to 472$ across layers $8$, $16$, $29$). \textbf{Bottom (Qwen3-VL-8B):} \emph{Early, non-monotonic mixing.} Mixed centroids follow a non-monotonic trajectory ($464 \to 61 \to 128$ across layers $9$, $18$, $27$).}
    \label{fig:umap_modality_gap}
\end{figure}

Prior work characterizes the \emph{modality gap} globally: image and
text embeddings occupy separated regions despite being trained under a
shared objective~\citep{liang2022mind}. LENS instead asks where this gap
is locally bridged within the VLM residual stream. We classify an MFA
component as \emph{mixed} when neither modality constitutes at least a
fraction $\tau$ of its hard-assigned activations, where assignments are
determined by maximum responsibility. We use $\tau{=}0.8$; a threshold
ablation is reported in Appendix~\ref{app:tau-sensitivity}.
\begin{itemize}
    \item \emph{Architecture-dependent fusion trajectories.}
Figure~\ref{fig:umap_modality_gap} shows UMAP projections
\citep{mcinnes2020umapuniformmanifoldapproximation} of the MFA
centroids across depth. LLaVA exhibits a monotonic increase in mixed components at later layers: their number rises from $0$ at layer~$8$ to $55$
at layer~$16$ and $472$ at layer~$29$ ($0\%$, $0.6\%$, and $5.7\%$,
respectively). Visual and textual neighborhoods are therefore initially
separated and become progressively interleaved toward the output.
In contrast, Qwen3-VL exhibits a distinct, non-monotonic trajectory. It contains
$464$ mixed components at layer~$9$, only $61$ at layer~$18$, and $128$
at layer~$27$ ($5.7\%$, $0.7\%$, and $1.5\%$). This early mixing,
partial re-segregation, and late recombination is consistent with
DeepStack's injection of multi-level visual features into early LLM
layers~\citep{meng2024deepstack,yang2025qwen3technicalreport}. Thus, the prevalence of mixed neighborhoods need not increase monotonically with depth and differs substantially between the two architectures.

\item\emph{Semantic content of mixed neighborhoods.}
The selected mixed components exhibit semantic coherence across modalities: their highest-responsibility image patches and text contexts refer to related concepts.
Figure~\ref{fig:mixed_components_main} shows a LLaVA component
associated with rockets and space launches and a Qwen3-VL component
associated with polar bears. The examples are selected by the automated
procedure described in Appendix~\ref{app:mixed-examples}, which also
provides additional cases.

\end{itemize}
Together, these results illustrate the advantage of a local geometric
decomposition: rather than reducing cross-modal fusion to a single
global separation measure, LENS identifies neighborhoods containing representations from both modalities and tracks how their prevalence changes across model depth.

\subsection{Cross-Modal Activation Steering}
\label{sec:exp_steering}

The geometry analysis identifies neighborhoods that contain both visual
and textual evidence, but does not establish whether they influence model
behavior. We test whether interventions toward MFA centroids causally influence model outputs by crossing the intervention scope--text, vision, or both token types--with
the modality dominance of the target component, yielding six
scope$\to$target conditions. Same-modality conditions test direct
concept control, while cross-modal conditions test whether a concept
localized primarily through one modality can be accessed by intervening
on the other.

\paragraph{Setup.}
For each model and target layer, we select $100$ labeled components ($50$ text-dominated and $50$ vision-dominated) and steer prefill activations at the selected visual and/or textual positions toward their MFA centroids via interpolation (Equation~\ref{eq:steering}), sweeping $\alpha$ over $15$ values in $[0.15,0.8]$. Each condition produces eight continuations; the image-only scope uses the prompt \textit{''Describe the image:''}, while the text and both scopes use \textit{''I think that''}, with vision-inclusive scopes additionally receiving a neutral gray image. A Qwen2.5-32B-Instruct-AWQ judge \citep{qwen2025qwen25technicalreport} scores concept alignment $\mathrm{CS}\in\{0,1,2\}$ and fluency $\mathrm{FL}\in\{0,1,2\}$. To ensure deterministic and reproducible evaluations, the judge utilizes greedy decoding with a temperature of 0 and a maximum generation limit of 150 tokens. We average $\mathrm{CS}$ and $\mathrm{FL}$ independently over the eight generations and report the harmonic mean of these two means, following AxBench \citep{wu2025axbench}. To measure attainable steering performance, we report each component’s best score over the evaluated $\alpha$ sweep and then take the median across components. We apply the same per-component selection protocol to all computational methods.

\begin{figure}[t]
    \centering
    \begin{minipage}{0.48\columnwidth}
        \centering
        \includegraphics[width=\linewidth]{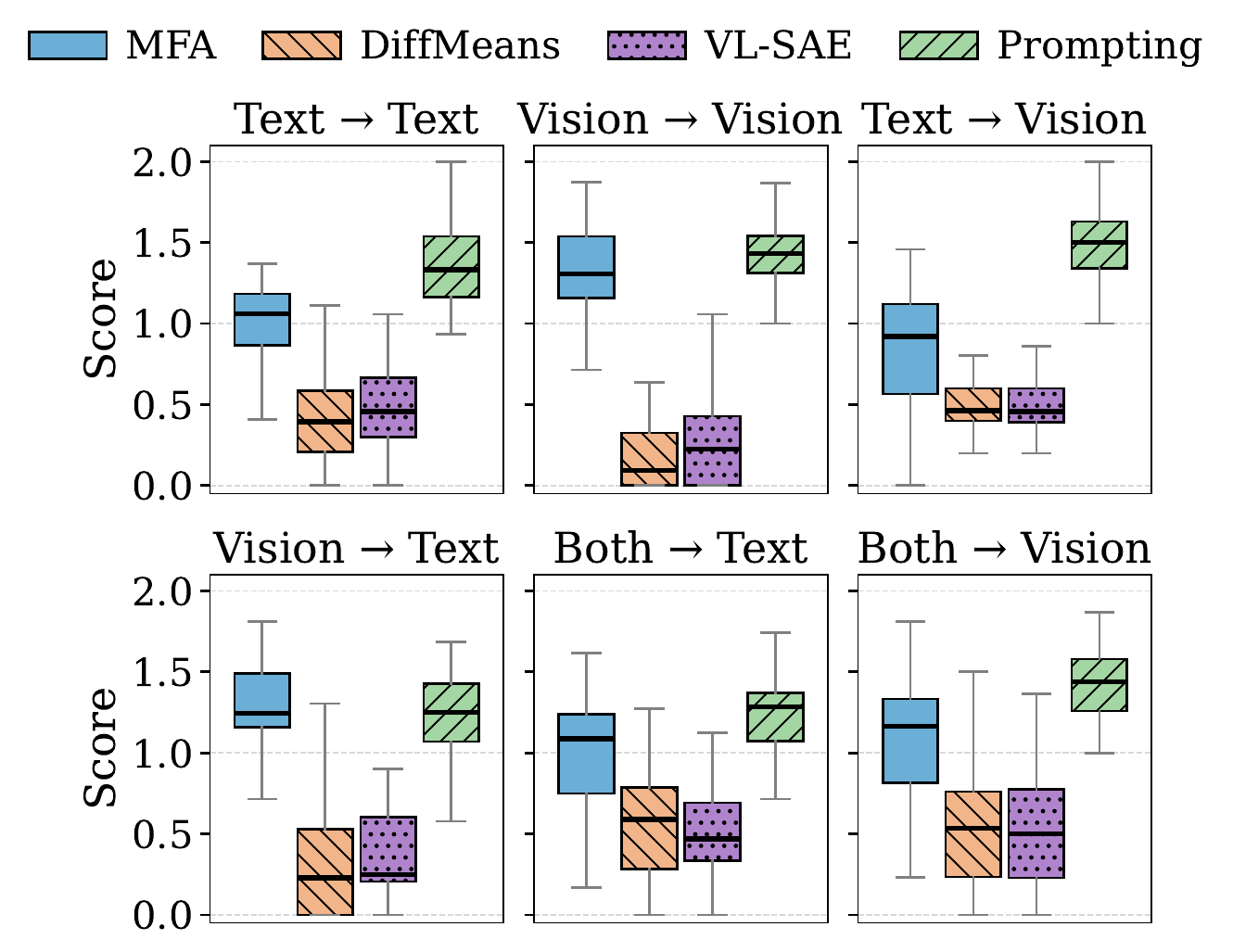}
        \centering \small (a) LLaVA (Layer 8)
    \end{minipage}\hfill
    \begin{minipage}{0.48\columnwidth}
        \centering
        \includegraphics[width=\linewidth]{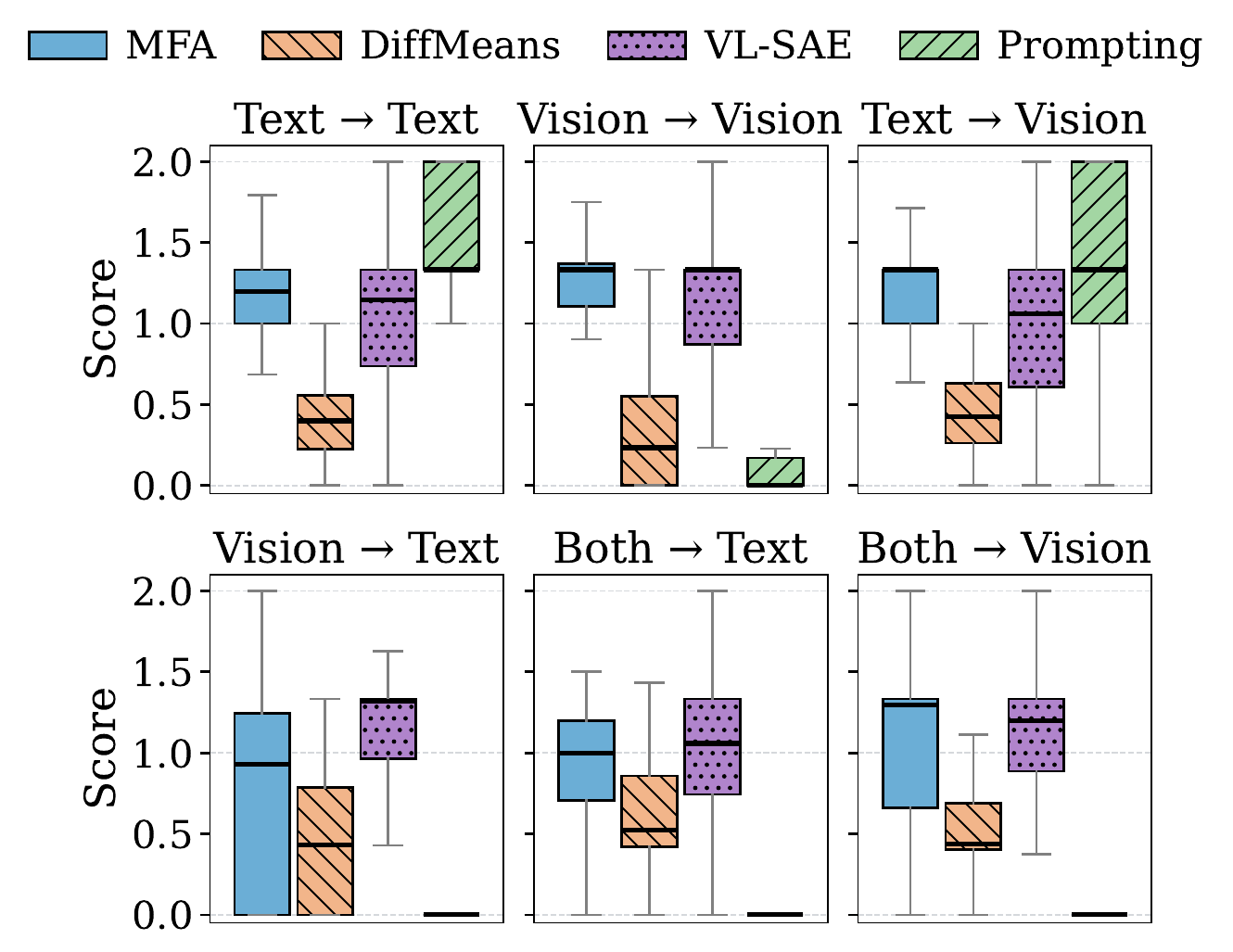}
        \centering \small (b) Qwen3-VL (Layer 9)
    \end{minipage}

    \vspace{0.2cm}
    \begin{minipage}{0.48\columnwidth}
        \centering
        \includegraphics[width=\linewidth]{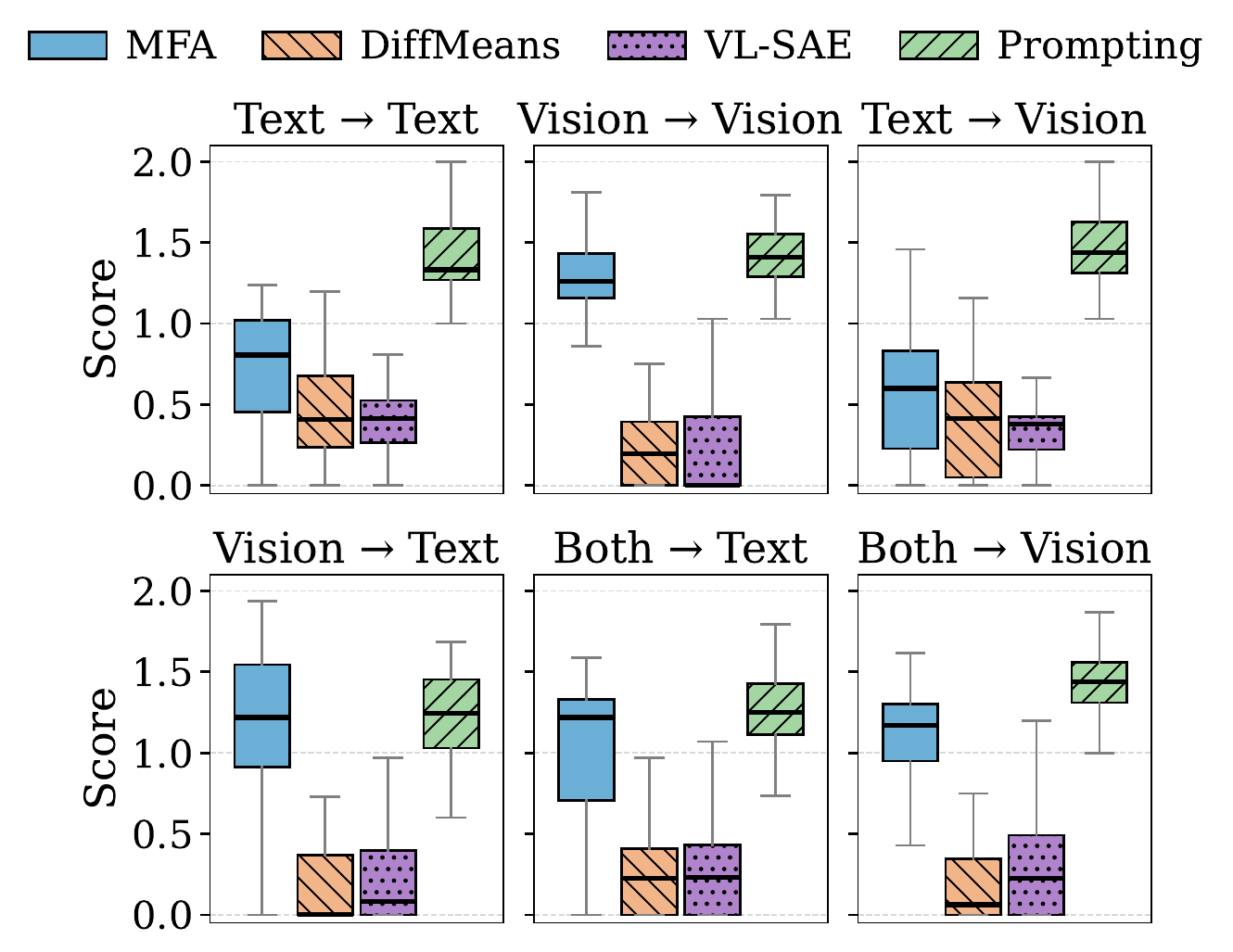}
        \centering \small (c) LLaVA (Layer 16)
    \end{minipage}\hfill
    \begin{minipage}{0.48\columnwidth}
        \centering
        \includegraphics[width=\linewidth]{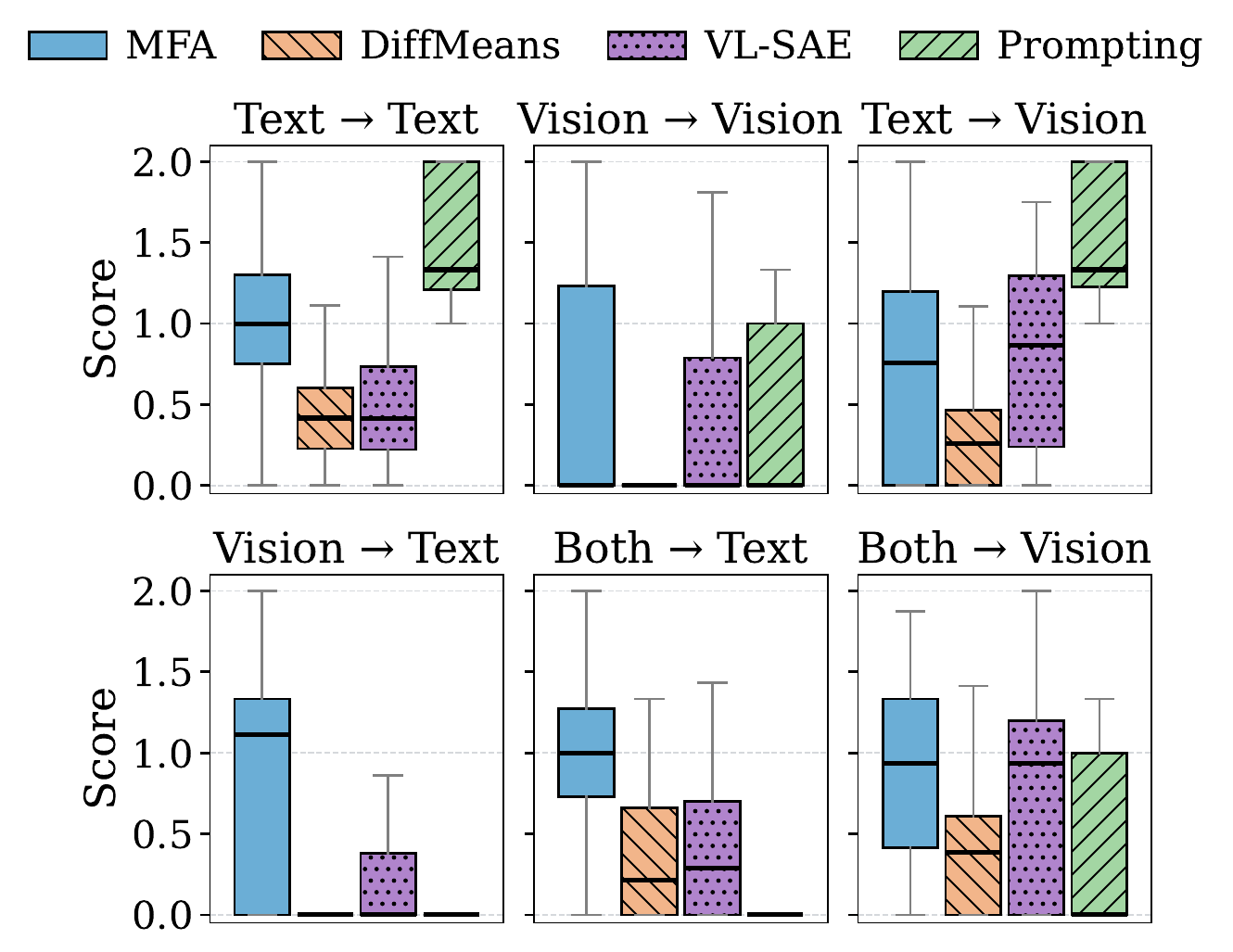}
        \centering \small (d) Qwen3-VL (Layer 18)
    \end{minipage}

    \caption{\textbf{MFA centroids support within- and cross-modal control.} Distributions show each component’s best-$\alpha$ harmonic score across six intervention-scope-to-target-modality conditions at early and middle layers. Boxes show the interquartile range, horizontal lines indicate the median,  and whiskers extending to 1.5× the interquartile range. MFA generally outperforms the computational baselines and transfers concepts across modalities; prompting is shown only as a non-interventional reference.}
    \label{fig:steering-low-mid}
\end{figure}


\paragraph{Baselines.}
We compare against two computational baselines and a prompting
reference. (1) \textbf{DiffMeans} \citep{rimsky-etal-2024-steering}
uses additive steering, $h'=h+\alpha v$, where $v$ is computed from
$72$ positive and $72$ neutral sentence templates instantiated with
each MFA concept label; the same templates are used across concepts.
We sweep $14$ values of $\alpha\in[0.4,100]$.
(2) \textbf{VL-SAE} adapts \citet{shen2026vlsae} by extracting unit
directions from VL-SAEs trained on the same residual-stream layers as
MFA. VL-SAE is evaluated on its own learned concepts, selected using
the same filtering criteria as the MFA concepts. Extraction details
appear in Appendix~\ref{app:vl_sae_extraction}. We apply
$h'=h+\alpha\lVert h\rVert\mathbf{v}_{\mathrm{unit}}$ over $21$ values
of $\alpha\in[0,100]$. For both baselines, we evaluate
both signs and retain the better-performing direction.
(3) \textbf{Prompting} directly inserts the target label. Text scope steering
uses \textit{In the following sentences I will discuss the concept:
$\{concept\}$. I think that:''}; vision scopes uses \textit{Focusing on the concept of $\{concept\}$, describe the image:''}. Prompting is a non-interventional reference and cannot independently mask token types.

\paragraph{Results.} Figure~\ref{fig:steering-low-mid} presents per model the early- and middle-layer
steering results; late-layer results appear in
Appendix~\ref{app:late-steering}.  The figure shows the distribution of the harmonic-mean values  across six intervention scope-to-target-modality conditions, with boxes indicating the interquartile range of the harmonic-mean values of the concepts, horizontal lines the median,  and whiskers extending to 1.5× the interquartile range.

\textbf{(i) MFA achieves higher best-sweep scores than the global-direction baselines in most conditions.} MFA centroid steering exceeds DiffMeans and VL-SAE across most
conditions, with particularly large gains for vision-targeted and
mixed-scope interventions. At LLaVA layer~$8$, MFA achieves
$\textit{Vis}{\to}\textit{Vis}{=}1.307$, compared with $0.094$ for
DiffMeans and $0.227$ for VL-SAE--5.7 times the VL-SAE score in this condition. Representative outputs are reported in
Figure~\ref{tab:steering-output-examples}.

\textbf{(ii) Steering effects transfer across modalities.} At LLaVA layer~$8$, steering vision tokens toward a text-dominated centroid obtains $\textit{Vis}{\to}\textit{Text}{=}1.244$, close to the same-modality $\textit{Vis}{\to}\textit{Vis}$ result. Conversely, steering text tokens toward a vision-dominated centroid obtains $\textit{Text}{\to}\textit{Vis}{=}0.921$. Thus, concepts localized primarily in one modality remain causally accessible through interventions on the other.

\textbf{(iii) Relative steering performance shifts with depth.}
At the early layers of both models, vision-dominated components
outperform text-dominated components in five of six conditions, with the
highest observed score at Qwen3-VL layer~$9$
($\textit{Vis}{\to}\textit{Vis}{=}1.333$). At the middle layers, this
pattern reverses in five of six conditions. The mechanism underlying this depth-dependent shift remains an open question.

\paragraph{Human evaluation.}
We evaluated 80 outputs, balanced across MFA, prompting, VL-SAE, and
DiffMeans, using human ratings of concept alignment and fluency on a
three-point scale. We averaged each rating across annotators and report
their harmonic mean. MFA achieved the highest overall median score
($1.64$), ahead of prompting ($1.43$), VL-SAE ($0.31$), and DiffMeans
($0.30$). Prompting slightly led MFA on LLaVA-1.5 ($1.84$ vs.\ $1.81$),
whereas MFA led on Qwen3-VL-8B ($1.43$ vs.\ $0.91$).

Across 75 distinct steering configurations after merging duplicates\footnote{Of the 80 evaluated outputs, nine shared identical steering configurations with another output (in four groups); we aggregated each group into a single item, yielding 75 distinct items for this correlation analysis.},
human and VLM-judge scores were correlated
(Spearman's $\rho=0.727$; Pearson's $r=0.747$), supporting the automated
evaluation while motivating independent human validation. Full
protocol and survey instructions are provided in
Appendix~\ref{app:human_eval}.

\begin{figure}[t]
    \centering
    \includegraphics[
        width=0.47\textwidth
    ]{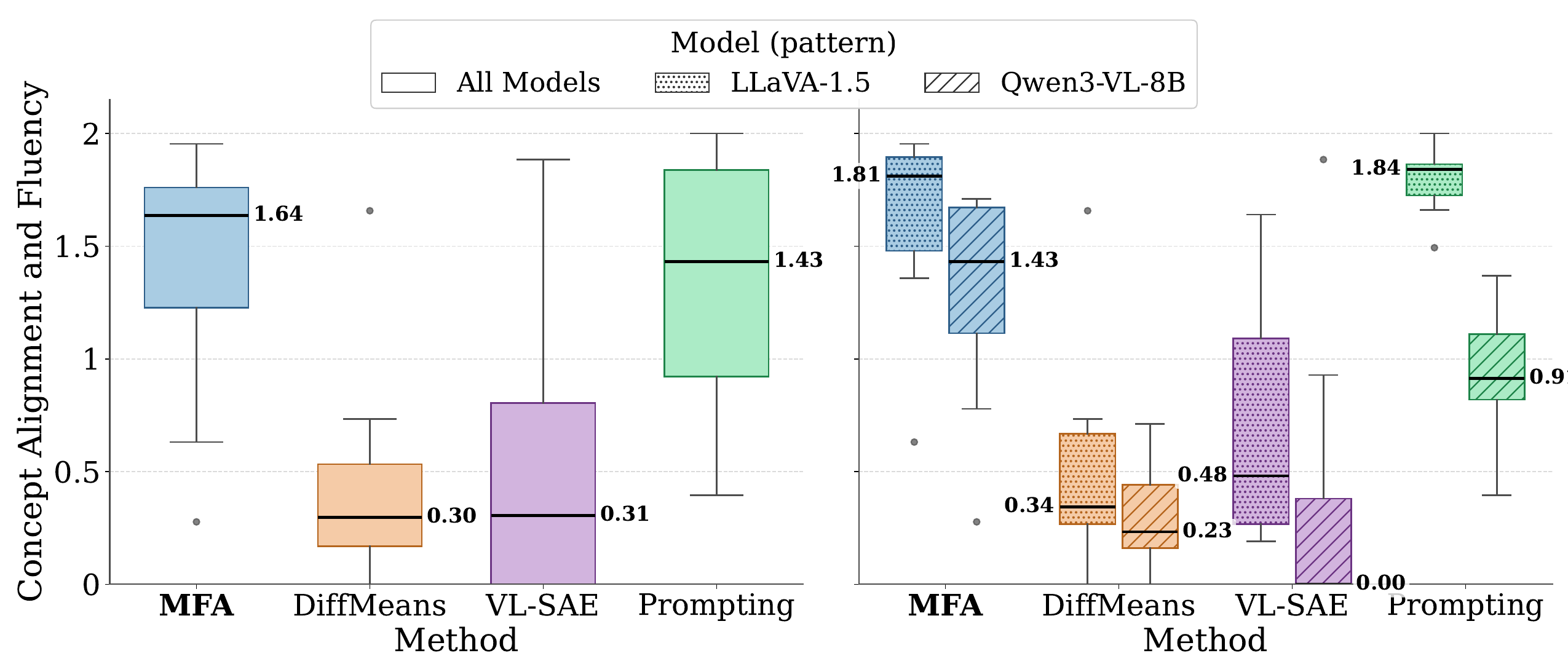}
    \caption{
        \textbf{Human evaluation of steering outputs.} Harmonic-mean scores combining concept alignment and fluency, shown overall and separately for each model. Boxes show the interquartile range of the harmonic-mean values, horizontal lines the median, and whiskers extending to 1.5× the interquartile range. MFA has the highest overall median, although prompting slightly leads on LLaVA-1.5 and MFA leads on Qwen3-VL-8B.
    }
    \label{fig:human_eval}
\end{figure}

\subsection{Cross-Modal Retrieval in Coefficient Space}
\label{sec:exp_retrieval}

\begin{figure}[t]
    \centering
    \includegraphics[width=0.47\textwidth]{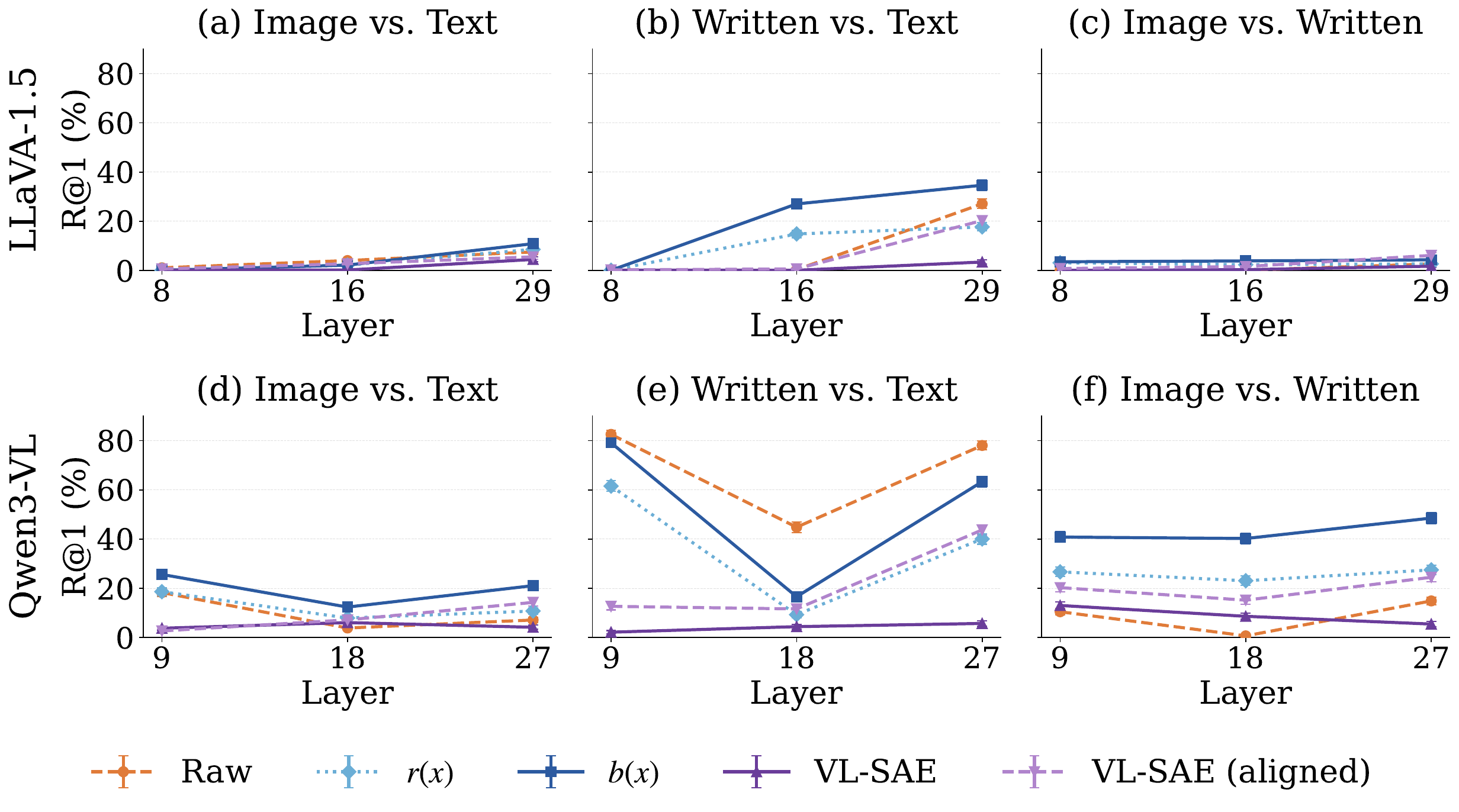}
    \caption{\textbf{MFA coordinates expose cross-modal alignment.}
    Bidirectional R@1 across layers for LLaVA-1.5 (top) and Qwen3-VL
    (bottom). We compare raw residual-stream activations, MFA
    responsibilities $\mathbf{r}(\mathbf{x})$, full MFA coefficients
    $\mathbf{b}(\mathbf{x})$, and standard and aligned VL-SAE
    representations. Error bars show 95\% bootstrap confidence
    intervals.}
    \label{fig:main_retrieval}
\end{figure}

Steering establishes that MFA neighborhoods are causally accessible; we
next test whether their coordinates encode shared semantic content across
modalities. Unlike contrastive representations, the MFA coefficient
space is learned without additional contrastive or task-specific alignment training.

\paragraph{Setup.}
Following \citet{vansprang2026contentdifferentanswerscrossmodal}, we use
ImageNet \citep{ILSVRC15_imagenet}, whose standardized $1{,}000$-class
label space enables controlled cross-modal comparison. For each class, we
construct a natural image, a rendered class label, and a textual class
label. Image--Text tests object--language alignment; Written--Text
isolates identical lexical content across pixels and tokens; and
Image--Written tests alignment between natural images and rendered class
names. We mean-pool token representations over modality-relevant
positions, compute an $N{\times}N$ cosine-similarity matrix for each
modality pair, model, and layer, and report bidirectional R@1 with 95\%
confidence intervals from $10{,}000$ bootstrap resamples.

\paragraph{Results.}
Figure~\ref{fig:main_retrieval} compares raw activations, MFA
responsibilities, full MFA coefficients, VL-SAE, and an aligned VL-SAE
representation extracted after its first cross-modal alignment encoder.

\textbf{(i) Local coordinates expose correspondences missed by global
representations.}
At Qwen3-VL layer~$27$, Image--Written R@1 increases from $14.9\%$ for
raw activations to $48.6\%$ for $\mathbf{b}(\mathbf{x})$. At LLaVA
layer~$29$, Image--Text retrieval similarly rises from $27.1\%$ to
$34.7\%$. Thus, the MFA coefficient space reveals semantic
correspondences that are only weakly expressed by global
residual-stream similarity.

\textbf{(ii) The advantage is largest at intermediate depths.}
In several middle-layer conditions, raw-activation R@1 falls to
$0.4\%$--$0.7\%$, while $\mathbf{b}(\mathbf{x})$ remains above $27\%$.
For reference, random R@1 with $1{,}000$ candidates is $0.1\%$.
Cross-modal content can therefore remain locally aligned even when
global vector angles provide little retrieval signal.

\textbf{(iii) The coefficient space preserves model-specific
differences.}
LLaVA remains below $5\%$ on Image--Written retrieval, whereas Qwen3-VL
performs substantially better. LENS therefore exposes, rather than
artificially removes, architecture-specific differences in how visual,
written, and textual inputs are represented. Together with the steering
results, this shows that MFA neighborhoods are both causally useful and
semantically structured across modalities.
\section{Discussion}
\label{sec:discussion}

\paragraph{Summary.} LENS identifies local, low-rank neighborhoods in fused VLM residual streams that are architecture-dependent, causally effective, and cross-modally aligned. MFA reveals progressive late fusion in LLaVA and early, non-monotonic fusion in Qwen3-VL, enables stronger within- and cross-modal steering than difference-in-means and VL-SAE across most conditions, and improves cross-modal retrieval over raw activations. Human evaluation supports the steering results, although the relative performance of MFA and prompting varies by architecture. Together, these findings establish local geometric neighborhoods as interpretable and steerable units
of VLM analysis.

\paragraph{Limitations.} Our study is limited to two open VLMs, CC3M activations, a controlled
ImageNet retrieval setting, and model-based component labels and
steering scores. Human evaluation covers only 80 outputs, late-layer
steering is less reliable, and MFA does not identify the circuits that
produce the recovered neighborhoods.

\paragraph{Future work.} Future work should test LENS across more architectures, datasets, and tasks. Tracing how attention and MLP layers form, merge, and separate MFA neighborhoods could connect geometric structure to mechanism. Comparing image-only, text-only, and paired activations could further quantify concept-specific modality gaps and their consistency for safety-relevant concepts.

\FloatBarrier

\bibliography{main}

\appendix
\clearpage
\appendix
\section{Implementation and Evaluation Details}
\label{app:implementation}

\subsection{Hardware and Software Specification}
\label{app:hardware_and_software}
All experiments were run on a single NVIDIA RTX 6000 GPU with 48\,GB
of VRAM and 80\,GB of system RAM. The complete software environment,
including package versions, is provided in the anonymized repository
and supplementary code archive.

\subsection{VL-SAE Direction Extraction and Bias Cancellation}
\label{app:vl_sae_extraction}
To isolate the pure concept direction $v_{unit}$ from the pretrained VL-SAE \citep{shen2026vlsae}, we perform a one-hot passthrough at the target concept dimension in the latent space. The VL-SAE utilizes a cascaded decoding architecture consisting of a modality-specific decoder and an auxiliary autoencoder decoder, both formalized as linear transformations ($y = Wx + b$). Consequently, a standard one-hot passthrough yields the target column weight plus the accumulated biases from both stages ($W_{:, i} + b_1 + b_2$).
To prevent these global reconstruction biases from skewing the geometric intervention, we execute a bias-cancellation protocol. We pass a zero-vector through the cascaded decoders to isolate the pure bias terms ($b_1 + b_2$) and subtract this from the one-hot output:
\begin{equation}
v_{pure} = (W_{:, i} + b_1 + b_2) - (b_1 + b_2) = W_{:, i}
\end{equation}
The resulting pure weight vector is then $L_2$-normalized to yield the unit-norm steering direction:
\begin{equation}
v_{unit} = \frac{v_{pure}}{\lVert v_{pure} \rVert_2}
\end{equation}

\subsection{Automated Labeling and Component Selection}
\label{app:automated_labeling}

\subsubsection{Automated Labeling Prompt.}
\label{app:prompt_labeling}

To synthesize concise, natural-language descriptions for the extracted MFA components, we provide the external judge VLM (Qwen2.5-VL-7B-Instruct) with the following system prompt. The prompt interleaves text snippets and image crops, along with their respective responsibility scores, enforcing a strict parsing format for the final output.

\vspace{-0.1cm}

\begin{tcolorbox}[
    breakable,
    colframe=black,
    colback=white,
    colbacktitle=black,
    coltitle=white,
    title=\textbf{Prompt 1: Multimodal Centroid Description},
    fontupper=\small\sffamily,
    boxrule=0.5pt,
    arc=2pt,
    left=6pt, right=6pt, top=6pt, bottom=6pt,
    width=\linewidth
]
You are a meticulous AI researcher conducting an important investigation into multimodal latent space representations. You will be given examples that represent the most strongly activating stimuli for a specific neural network component.

\vspace{0.5em}
\textbf{Examples come in two forms:}
\begin{enumerate}
    \item \textbf{Text snippets:} The target token is highlighted between \texttt{<<} \texttt{>>} delimiters within its surrounding context.
    \item \textbf{Images:} A visual crop showing the content at the activation site.
\end{enumerate}

\textbf{Guidelines:}
\begin{itemize}
    \item Analyze the examples and produce one concise natural-language interpretation that captures the shared pattern present across the stimuli.
    \item Focus on describing the semantic, visual, syntactic, or conceptual pattern uniting the examples.
    \item Each example includes a score between 0 and 1. Higher values indicate that this example is more mathematically important for the concept.
    \item If the examples are uninformative or appear as noise, give the most concise possible summary of the pattern you can infer.
    \item Do not repeat or reference the marker tokens (\texttt{<<} \texttt{>>}) in your interpretation.
    \item Do not list multiple possible interpretations or speculate; provide a single, clear, crisp description.
    \item Keep your interpretation short, direct, and precise. Be as specific as possible so that the description is unambiguous.
\end{itemize}

\textbf{RESPONSE FORMAT (STRICT):}
\begin{itemize}
    \item You may optionally include a brief explanation first.
    \item The FINAL line of your response MUST be exactly of the form:\\
    \texttt{[interpretation]: <your one-sentence description>}
    \item That line MUST start with \texttt{"[interpretation]:"}, MUST be plain text (no markdown bullets, no code fences, no quotes around it), MUST appear exactly once, and MUST be the last line of your response.
    \item Do NOT output anything after that line.
\end{itemize}
\end{tcolorbox}

\subsubsection{Component Selection for Steering Evaluation.}
\label{app:component-selection}

To ensure the 100 components selected for causal steering (Section~\ref{sec:exp_steering}) represented genuine semantic concepts rather than narrow syntactic artifacts (e.g., single-token patterns), we utilized an LLM judge to rigorously filter the labeled candidate pool. The judge was provided with each component's identifier, its automated label (generated via the protocol in Appendix~\ref{app:automated_labeling}), and its modality distribution. It was then prompted to select exactly 50 text-dominant and 50 vision-dominant components based on the following strict criteria:

\vspace{-0.1cm}

\begin{tcolorbox}[
    breakable,
    colframe=black,
    colback=white,
    colbacktitle=black,
    coltitle=white,
    title=\textbf{Component Selection Prompt},
    fontupper=\small\sffamily,
    boxrule=0.5pt,
    arc=2pt,
    left=6pt, right=6pt, top=6pt, bottom=6pt,
    width=\linewidth
]
You are selecting MFA Gaussian components for a steering experiment on a multimodal VLM. The goal is to find components whose centroids, when used to interpolate hidden states, will visibly push model generation toward a specific concept. For each component, you are given: component\_id, label (description of what the component captures), and modality\_distribution (text\_ratio, image\_ratio). Select exactly $50$ text-dominant components (text\_ratio $\geq 0.8$) and $50$ vision-dominant components (image\_ratio $\geq 0.8$). Selection criteria, in priority order:
\begin{enumerate}
    \item Concrete and specific: the label names things that would appear as recognizable words in a 50-token text completion. "Medical procedures and surgical instruments" is good. "Various patterns and designs" is bad. "The images primarily feature a person" is bad - too vague.
    \item Testable: a judge reading the completion could determine whether the concept is present without ambiguity. "Sports teams and athletic events" is testable. "Abstract compositional structures" is not.
    \item Diverse: spread across different topics. Don't pick 10 variants of food-related components.
    \item Avoid single-token Gaussians: skip components whose label describes only a token pattern (e.g., "the word 'in' in various contexts").
\end{enumerate}
For each selected component, output: component\_id, modality ("text" or "vision"), label. Output as a JSON array. The list of components will be provided in the following messages.
\end{tcolorbox}

To guarantee a rigorous and balanced baseline comparison, the 100 native concepts sampled for the VL-SAE evaluation were filtered using a structurally identical LLM judge prompt. This ensures that the exact same standards for concreteness, diversity, and testability were enforced across both representation spaces prior to steering.

\subsection{Human Evaluation Protocol}
\label{app:human_eval}

We evaluated 80 generated outputs, with 20 outputs sampled from each method: MFA, DiffMeans, VL-SAE, and prompting. Forty annotators participated, and each annotator rated 20 randomly assigned outputs.
Assignments were balanced such that each output received ratings from 10 annotators, yielding 800 item--annotator evaluations.

For each concept--text pair, annotators independently rated concept alignment and generation fluency on the same three-point ordinal scale as the VLM judge, with scores in $\{0,1,2\}$. For each output, we first
averaged alignment and fluency separately across annotators. We then
combined the two mean ratings using their harmonic mean:
\[
H = \frac{2\bar{A}\bar{F}}{\bar{A}+\bar{F}},
\]
where $\bar{A}$ and $\bar{F}$ denote the mean alignment and fluency
ratings, respectively. We set $H=0$ when
$\bar{A}+\bar{F}=0$.

To assess agreement between human and automated evaluation, we matched
each output's human score to its VLM-judge score and computed Pearson
and Spearman correlations. Of the 80 outputs, nine shared a steering
configuration with another output, forming four duplicate groups. We
averaged scores within each duplicate group before computing the
correlations, yielding 75 distinct steering configurations. 

The complete instructions shown to annotators are reproduced below.

\vspace{-0.1cm}

\begin{tcolorbox}[
    breakable,
    colframe=black,
    colback=white,
    colbacktitle=black,
    coltitle=white,
    title=\textbf{Human Evaluation Instructions},
    fontupper=\small\sffamily,
    boxrule=0.5pt,
    arc=2pt,
    left=6pt, right=6pt, top=6pt, bottom=6pt,
    width=\linewidth
]

Thank you for helping with this survey.
There are 20 questions in the survey.

\vspace{0.5em}

\textbf{Important:}
For every concept--text pair, provide both ratings:
one for \textbf{Correlation} and one for \textbf{Fluency}.

\vspace{0.5em}

\textbf{Each question includes:}
\begin{itemize}
    \item \textbf{Concept:} A title or idea.
    \item \textbf{Text:} Text intended to represent that concept.
\end{itemize}

Rate how clearly and accurately the text represents the concept.

\vspace{0.5em}

\textbf{Correlation ratings:}
\begin{itemize}
    \item \textbf{Correlated (2):}
    The text clearly and accurately represents the concept.
    \item \textbf{Weakly correlated (1):}
    The text is somewhat related to the concept, but the connection is
    unclear or incomplete.
    \item \textbf{Uncorrelated (0):}
    The text does not meaningfully represent the concept.
\end{itemize}

\textbf{Fluency ratings:}
\begin{itemize}
    \item \textbf{Fluently readable (2):}
    The text is clear, natural, and easy to read.
    \item \textbf{Understandable (1):}
    The text can be understood, but it may be awkward, incomplete, or
    contain noticeable errors.
    \item \textbf{Unreadable (0):}
    The text is too nonsensical, fragmented, or unclear to understand.
\end{itemize}

\textbf{Keep in mind:}
\begin{itemize}
    \item If the text is cut off, do not automatically assign it a poor
    rating.
    \item If the text is written in a language you do not understand,
    you may rate it as uncorrelated.
\end{itemize}

\end{tcolorbox}
\section{Robustness and Sensitivity Analyses}
\label{app:robustness}

To assess whether our conclusions depend on modeling choices or
intervention depth, we examine sensitivity to MFA capacity $K$, local
rank $q$, the modality-purity threshold $\tau$, and the choice of
steering layer. Together, these analyses test the robustness of both
the geometric and causal findings presented in the main text.

\subsection{MFA Hyperparameter Sensitivity}
\label{app:ablation}
We assess the sensitivity of LENS to its two principal hyperparameters: the number of mixture components $K$ and the per-component latent rank $q$ (Section~\ref{sec:exp_setup}). Due to the computational overhead of refitting and evaluating multiple MFA variants, we conduct this comprehensive sweep on LLaVA-1.5, holding all other baseline settings fixed. 

We re-evaluate the two core claims upon which the main text relies: the geometric structure of cross-modal fusion (Section~\ref{sec:exp_geometry}) and the causal efficacy of the resulting components (Section~\ref{sec:exp_steering}). 

\begin{figure}[h]
    \centering
    \includegraphics[width=0.48\textwidth]{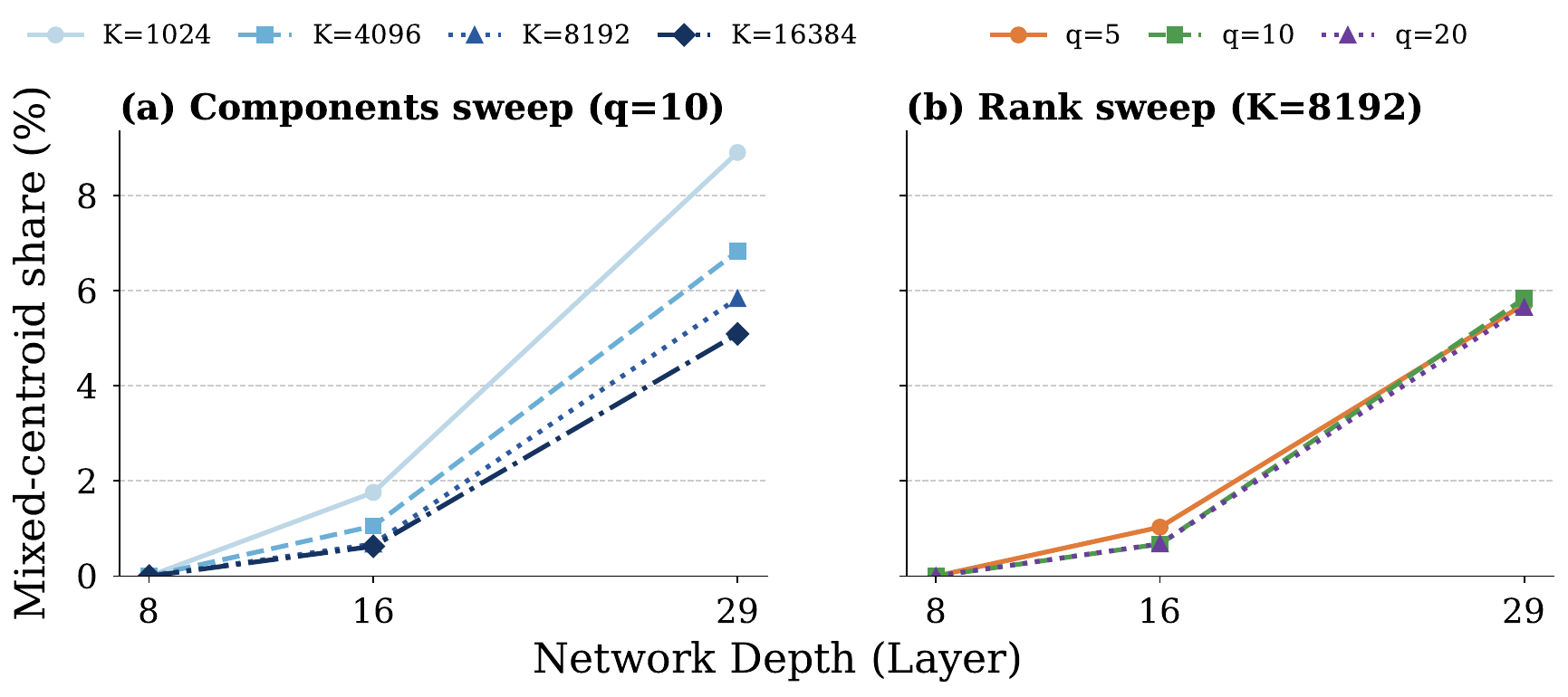}
    \caption{\textbf{Sensitivity of cross-modal fusion trajectories to MFA hyperparameters.}
    Mixed-centroid share across LLaVA-1.5 layers under (a) varying component counts $K$ and (b) varying local ranks $q$. The monotonic late-fusion trajectory is preserved across all settings.}
    \label{fig:abb_hyperparameter}
\end{figure}

To validate the geometric findings, we refit the variants across early, middle, and late layers, with results presented in Figure~\ref{fig:abb_hyperparameter}. Because models with higher capacity (larger $K$) naturally contain a larger volume of centroids, comparing raw counts is misleading. Therefore, we report the \emph{mixed-centroid share} (the percentage of total components classified as mixed) to properly normalize across the sweeps.  As demonstrated in the figure, while smaller mixture capacities ($K$) yield a higher percentage of mixed components due to the broader geometric volume of each local neighborhood, the fundamental cross-modal trajectory—a monotonic, late-binding curve—remains entirely invariant across all tested capacities and local ranks. 

To validate the causal findings, we refit the variants at a representative middle fusion layer (Layer 16) and measure the best-$\alpha$ steering score, detailed in Table~\ref{tab:causal_ablation}. Across three representative intervention scopes, the LENS framework consistently outperforms both the DiffMeans and VL-SAE baselines regardless of the mixture capacity ($K$) or the local latent rank ($q$). This confirms that the precise steerability of the discovered components is a robust property of the MFA formulation itself, rather than an artifact of hyperparameter configuration.

\vspace{-0.3cm}

\begin{table}[h]
    \centering
    \caption{\textbf{Causal steering hyperparameter ablation (LLaVA-1.5, Layer 16).} Median best-$\alpha$ harmonic scores across 50 components per condition. MFA outperforms both baselines across all evaluated values of $K$ and $q$; the main-text setting is shown in bold.}
    \begin{tabular}{lccc}
        \toprule
        \textbf{Method / Variant} & \textbf{Vis $\rightarrow$ Vis} & \textbf{Text $\rightarrow$ Text} & \textbf{Text $\rightarrow$ Vis} \\
        \midrule
        \multicolumn{4}{l}{\textit{Baselines}} \\
        DiffMeans & 0.194 & 0.409 & 0.413 \\
        VL-SAE & 0.000 & 0.416 & 0.378 \\
        \midrule
        \multicolumn{4}{l}{\textit{MFA Capacity Sweep ($q=10$)}} \\
        MFA ($K=1024$) & 1.238 & 0.645 & 0.742 \\
        MFA ($K=4096$) & 1.200 & 0.726 & 0.692 \\
        \textbf{MFA ($K=8192$)} & \textbf{1.261} & \textbf{0.806} & \textbf{0.600} \\
        MFA ($K=16384$) & 1.197 & 0.917 & 0.562 \\
        \midrule
        \multicolumn{4}{l}{\textit{MFA Local Rank Sweep ($K=8192$)}} \\
        MFA ($q=5$) & 1.219 & 0.738 & 0.614 \\
        \textbf{MFA ($q=10$)} & \textbf{1.261} & \textbf{0.806} & \textbf{0.600} \\
        MFA ($q=20$) & 1.219 & 0.718 & 0.557 \\
        \bottomrule
    \end{tabular}
    \label{tab:causal_ablation}
\end{table}

\vspace{-0.3cm}

\subsection{Modality-Purity Threshold Sensitivity}
\label{app:tau-sensitivity}
The mixed-centroid counts reported in Section~\ref{sec:exp_geometry} depend on the modality-purity threshold $\tau$. A centroid is defined as \emph{mixed} when neither the visual nor the textual modality accounts for a fraction $\ge \tau$ of its hard-assigned activations. 

\begin{figure}[h]
    \centering
    \includegraphics[width=0.47\textwidth]{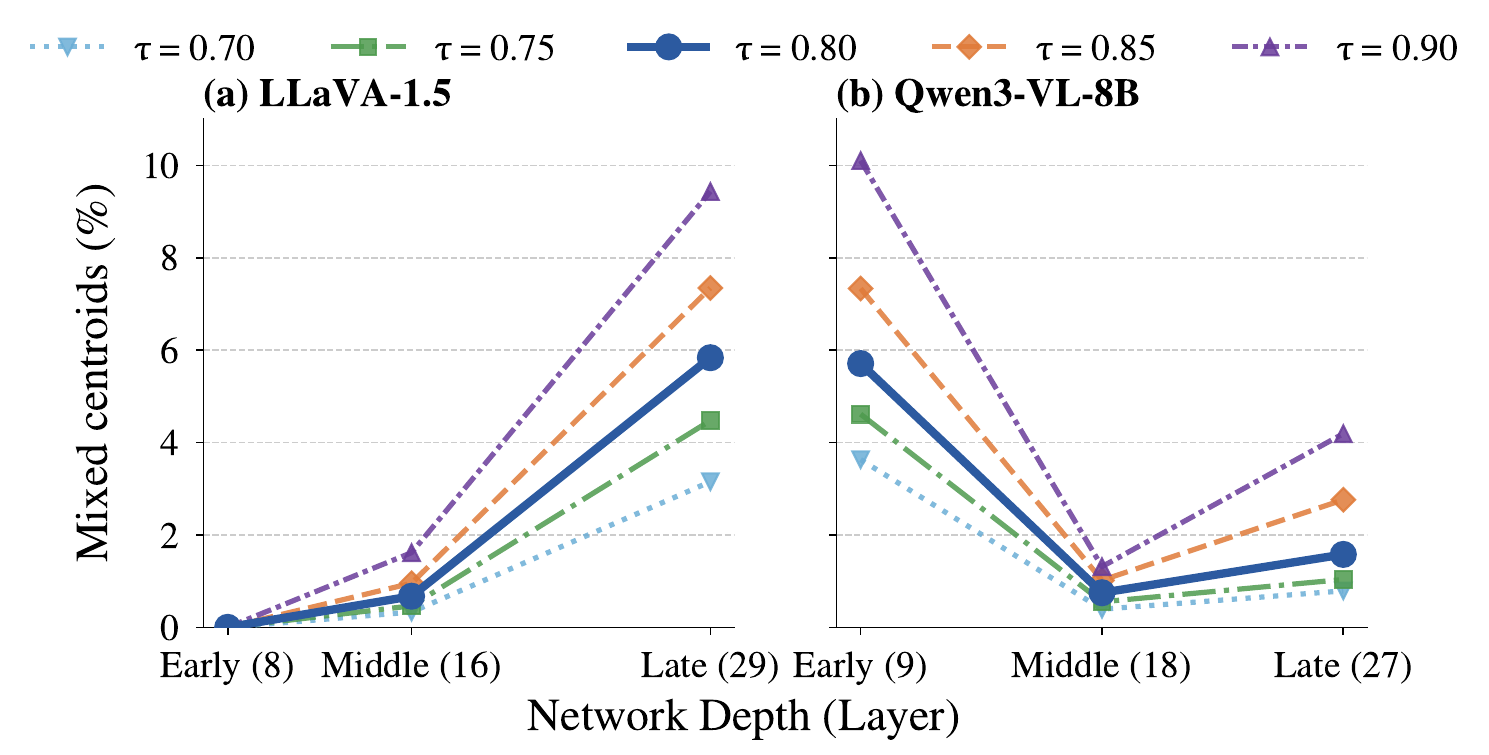}
    \caption{\textbf{Sensitivity of cross-modal fusion regimes to the modality-purity threshold $\tau$.} Mixed-centroid share across layers for (a) LLaVA-1.5 and (b) Qwen3-VL-8B. Although the magnitude changes with $\tau$, the architecture-specific trajectories remain stable; the main-text setting $\tau=0.80$ is highlighted.}
    \label{fig:abb_tau_sensitivity}
\end{figure}

Because raising $\tau$ sets a stricter bar for a component to be considered ``pure,'' it mathematically widens the bounds for a mixed classification. Consequently, higher values of $\tau$ yield higher absolute counts of mixed components. We sweep $\tau \in \{0.70, 0.75, 0.80, 0.85, 0.90\}$ to confirm that while the absolute magnitude scales with $\tau$, the underlying structural trajectories of the fusion regimes remain unchanged. As shown in Figure~\ref{fig:abb_tau_sensitivity}, LLaVA's monotonic, late-binding growth and Qwen3-VL's non-monotonic, early-and-deep binding are strictly preserved across all evaluated thresholds, confirming these patterns are robust architectural signatures rather than artifacts of our operating point ($\tau = 0.80$).

\vspace{-0.3cm}

\subsection{Late-Layer Steering and Rank Sensitivity}
\label{app:late-steering}
\begin{figure}[h]
  \centering
  \begin{subfigure}[b]{0.55\linewidth}
    \centering
    \includegraphics[width=\linewidth]{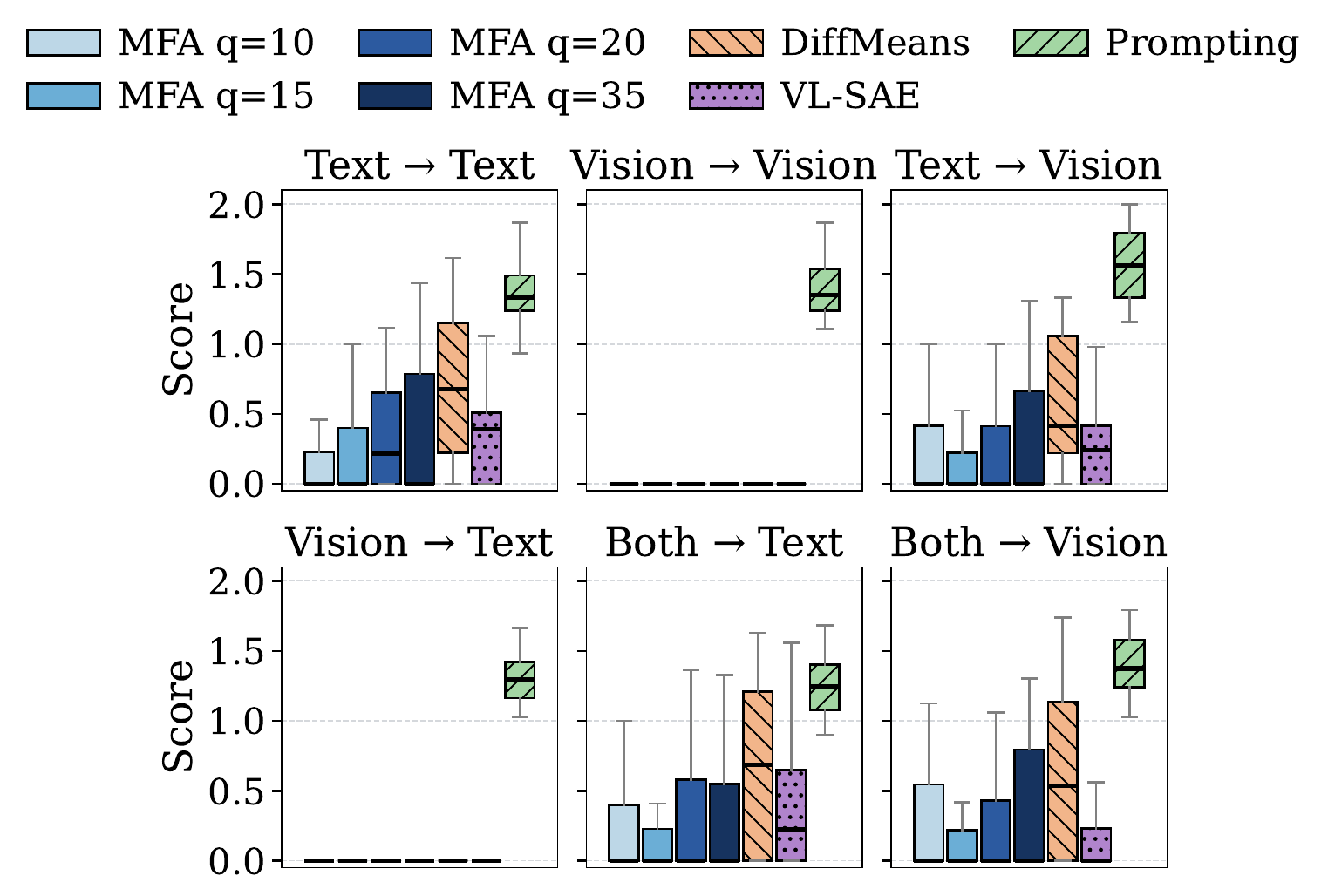}
    \caption{LLaVA-1.5 Layer 29}
    \label{fig:steering-late-llava}
  \end{subfigure}\hfill
  \begin{subfigure}[b]{0.45\linewidth}
    \centering
    \includegraphics[width=\linewidth]{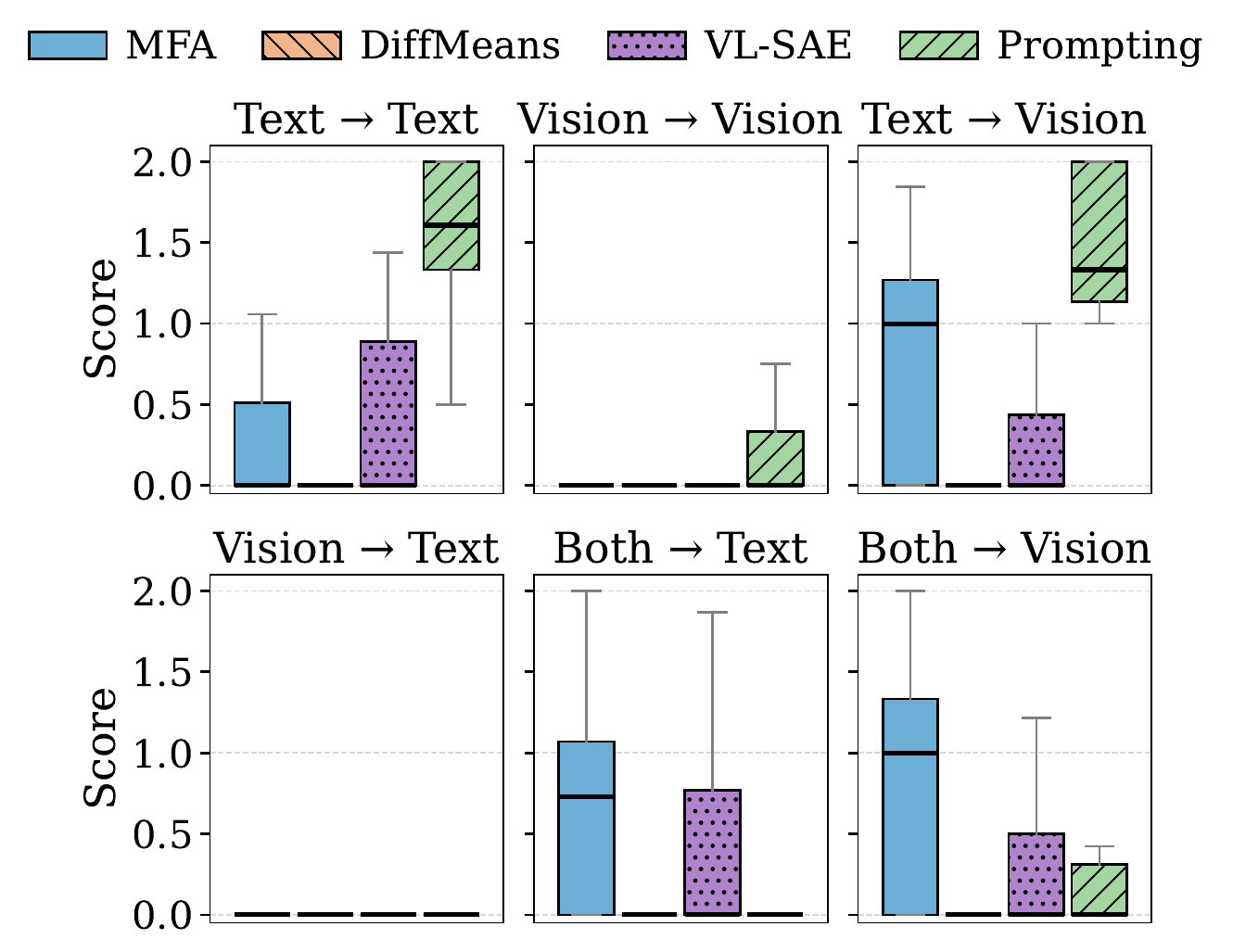}
    \caption{Qwen3-VL-8B Layer 27}
    \label{fig:steering-late-qwen}
  \end{subfigure}
  \caption{\textbf{Late-layer steering performance.}
    Distributions show each component’s best-$\alpha$ harmonic score across six intervention-scope-to-target-modality
conditions at late-layer: (a) LLaVA-1.5 Layer 29 across local ranks and (b) Qwen3-VL-8B Layer 27. Boxes show the interquartile range, horizontal lines indicate the median, and whiskers extending to 1.5x the interquartile range. Steering is generally weaker and less stable than at earlier layers.}
  \label{fig:steering-late}
\end{figure}

Following the early and middle layer evaluations (Section~\ref{sec:exp_steering}), Figure ~\ref{fig:steering-late} presents the late-layer steering results for Qwen3-VL-8b (layer 27) and an auxiliary sweep over the local latent rank q for LLaVA-1.5 (layer 29).
Empirically, MFA centroid steering at these depths is brittle. The steering scores drop compared to earlier layers, and DiffMeans often outperforms MFA in text-targeted scopes. Furthermore the LLaVA-1.5 sweep demonstrates that steering efficacy is sensitive to the local latent rank q. Notably, intervening strictly on vision tokens largely fails across both models. We hypothesize that visual information is already assimilated into the language model's native residual flow by these late stages, rendereing manipulation at the original visual token positions mostly ineffective. This degradation aligns with our geometric intuition: near the final vocabulary projection, interpolating towards an MFA centroid may not provide a sufficient geometric shift to redirect the output. Conversely, unconstrained additive steering (DiffMeans) pushes the activation more aggressively to force a vocabulary shift, bypassing the local geometry.

\vspace{-0.1cm}

\section{Additional Qualitative Results}
\label{app:qualitative}

\subsection{Additional Mixed-Component Examples}
\label{app:mixed-examples}

To ensure qualitative examples are free from manual cherry-picking, we
employ a strictly automated selection pipeline. We first establish a
candidate pool by filtering all components across the evaluated models
based on four objective criteria: valid automated judge parsing,
strictly multimodal hard-assignment distributions (enforced via a
maximum modality-purity threshold), a minimum count of hard-assigned
image patches to guarantee visual coherence, and substantive textual
density (requiring top-activating tokens to be content words rather
than prompt artifacts).

From this robust candidate pool, we select the final examples via a
deterministically seeded random permutation. Representative components
are featured in the main text, with additional examples presented in
Figure~\ref{app:mixed}.

\definecolor{tokbg}{RGB}{245,205,75}

\newcommand{\hltok}[1]{%
  {\setlength{\fboxsep}{0.6pt}%
   \colorbox{tokbg}{\textbf{#1}}}%
}

\newcommand{\tokstack}[1]{%
  {\footnotesize
   \renewcommand{\arraystretch}{0.88}%
   \begin{tabular}[t]{@{}l@{}}
     #1
   \end{tabular}}%
}

\newlength{\mixeddesignwidth}

\par
\vspace{4pt}

\noindent
\begin{minipage}{\columnwidth}
  \centering

  \setlength{\mixeddesignwidth}{1.28\columnwidth}

  \begin{adjustbox}{width=\columnwidth}
    \begin{minipage}{\mixeddesignwidth}
      \centering
      \small

      \mixedcomponent
        {Qwen3-VL-8B}
        {18}
        {4771}
        {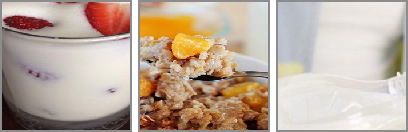}
        {%
          \tokstack{%
            bowl of por\hltok{ridge}\\
            overnight \hltok{oats} in a jar\\
            por\hltok{ridge} with berries\\
            bowl of por\hltok{ridge}\\
            yog\hltok{urt} with blueberries
          }%
        }
        {Breakfast items such as porridge, oatmeal, yogurt, and
         granola are strongly activating.}

      \vspace{-1pt}

      \mixedcomponent
        {Qwen3-VL-8B}
        {18}
        {6049}
        {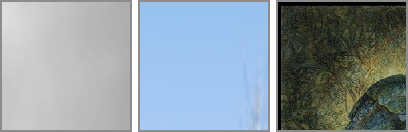}
        {%
          \tokstack{%
            p\hltok{ige}on sitting on\\
            the r\hltok{aven} by novelist\\
            flying \hltok{pige}ons in front\\
            p\hltok{ige}ons walking on\\
            a \hltok{crow} tucks a little
          }%
        }
        {Birds, particularly pigeons and ravens, across snowy,
         urban, and natural environments.}

      \vspace{-1pt}

      \mixedcomponent
        {Qwen3-VL-8B}
        {18}
        {1143}
        {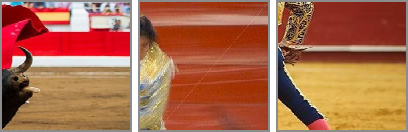}
        {%
          \tokstack{%
            roping a young \hltok{steer}\\
            during a bull \hltok{fight}\\
            distract the \hltok{bull} so\\
            passes by mat\hltok{ador}\\
            matador during a \hltok{bull}
          }%
        }
        {Bullfighting and rodeo scenes involving interactions
         between humans and bulls, including matadors, cowgirls,
         and assistants.}

      \vspace{-1pt}

      \mixedcomponent
        {Qwen3-VL-8B}
        {27}
        {6284}
        {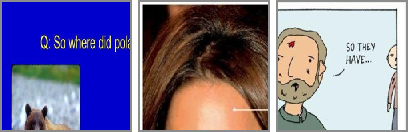}
        {%
          \tokstack{%
            \hltok{so} -- skinny jeans\\
            \hltok{so}aker hoses conserve\\
            \hltok{so}aring: video\\
            at the \hltok{so} called coast\\
            \hltok{so}, who do
          }%
        }
        {The token \texttt{so} appears in various contexts, often
         serving as a conjunction or intensifier.}

    \end{minipage}%
  \end{adjustbox}

  \captionsetup{
    font=footnotesize,
    skip=2pt
  }

  \captionof{figure}{\textbf{Additional mixed components.}
    Four components from multiple layers, shown with their
    highest-responsibility image patches, text contexts with the
    responsible token highlighted, and automated labels.}
  \label{app:mixed}
\end{minipage}

\par
\newpage

\subsection{Representative Steering Outputs}
\label{app:representative-steering}


\begin{figure}[h]
    \centering
 \includegraphics[width=\linewidth,trim=0 4.1cm 0 0, clip]{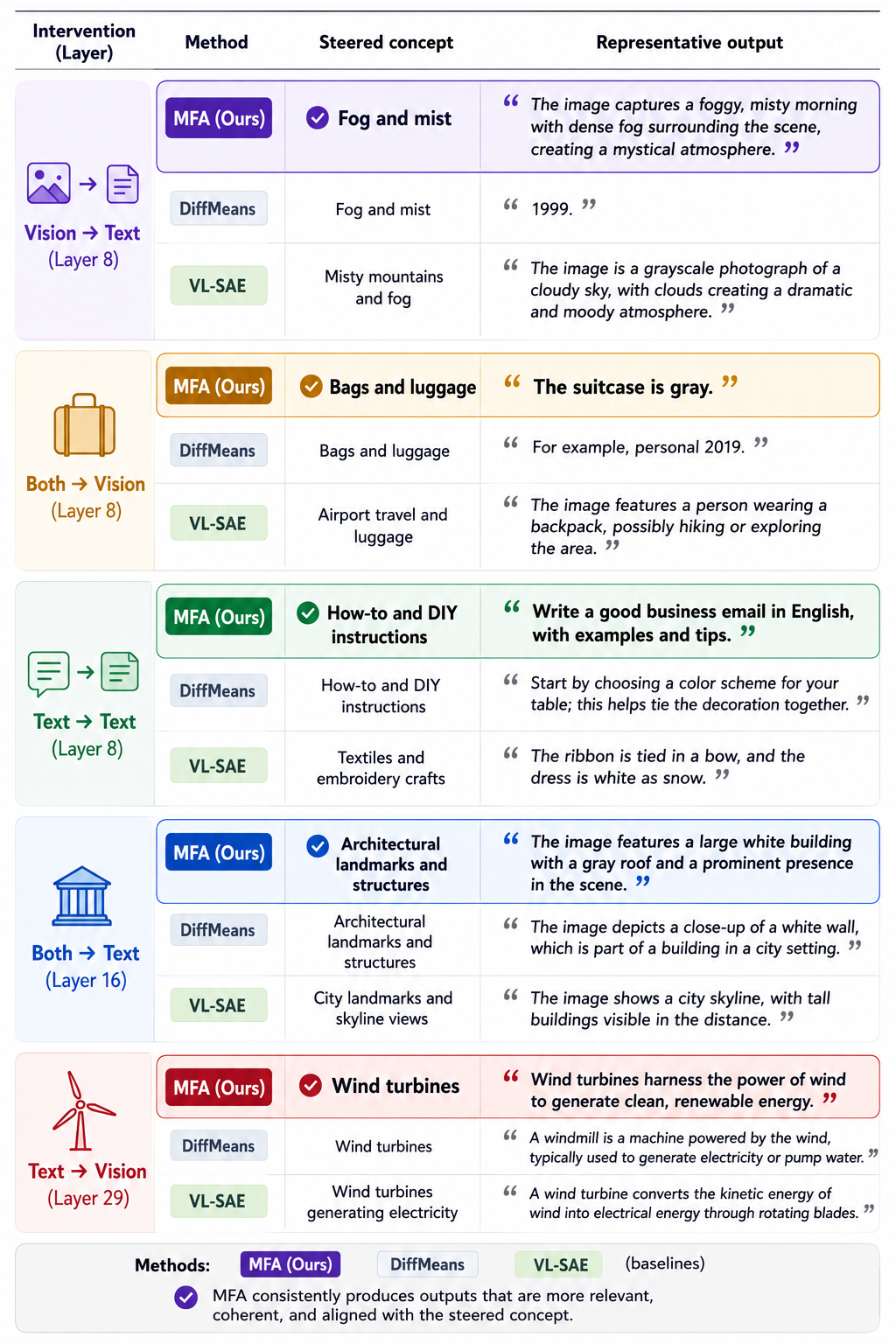}
    \caption{\textbf{Representative steering outputs for
    LLaVA-1.5-7B.} Each intervention condition includes one example
    from MFA, DiffMeans, and VL-SAE. Rows report the layer,
    intervention type, steering strength $\alpha$, harmonic score,
    target concept, and output.}
    \label{tab:steering-output-examples}
\end{figure}


\end{document}